\documentclass[preprint,3p,times,twocolumn]{elsarticle}

\usepackage{multirow}
\usepackage{colortbl}
\usepackage{hhline}
\usepackage{url}
\usepackage{amssymb}
\usepackage[noend]{algpseudocode}
\usepackage{algorithmicx,algorithm}
\usepackage[colorlinks,linkcolor=blue]{hyperref}

\usepackage{amssymb}

\begin{document}

\begin{frontmatter}



\title{FFR\_FD: Effective and Fast Detection of DeepFakes Based on Feature Point Defects}


\address[1]{Engineering Research Center of Cyberspace, Yunnan University, Kunming, China}
\address[2]{School of Software, Yunnan University, Kunming, China}
\address[3]{School of Cyber Science and Engineering, Wuhan University, Wuhan, China}

\author[1,2]{Gaojian Wang}
\author[1,2]{Qian Jiang}
\author[1,2]{Xin Jin}
\author[3,]{Xiaohui Cui}\fntext[label2]{Corresponding author. 
E-mail address: xcui@whu.edu.cn}

\begin{abstract}
The internet is filled with fake face images and videos synthesized by deep generative models. These realistic DeepFakes pose a challenge to determine the authenticity of multimedia content. As countermeasures, artifact-based detection methods suffer from insufficiently fine-grained features that lead to limited detection performance. DNN-based detection methods are not efficient enough, given that a DeepFake can be created easily by mobile apps and DNN-based models require high computational resources. For the first time, we show that DeepFake faces have fewer \textit{feature points} than real ones, especially in certain facial regions. Inspired by feature point detector-descriptors to extract discriminative features at the pixel level, we   propose the \textit{Fused Facial Region\_Feature Descriptor (FFR\_FD)} for effective and fast DeepFake detection. FFR\_FD is only a vector extracted from the face, and it can be constructed from any feature point detector-descriptors. We train a random forest classifier with FFR\_FD and conduct extensive experiments on six large-scale DeepFake datasets, whose results demonstrate that our method is superior to most state of the art DNN-based models.
\end{abstract}




\end{frontmatter}


\section{Introduction}
Misinformation has become ubiquitous, a consequence of the ability to easily create multimedia content and share it on social platforms. The remarkable development of deep generative models has further led to the synthesis of super-realistic images or videos. For example, the well-known DeepFake refers to swapping faces in a video through  generative adversarial networks (GANs) \cite{goodfellow2014generative} or autoencoders (AEs)  \cite{kingma2013auto}. Some mobile apps (e.g., Avatarify \cite{Avatarify} and Reface \cite{Reface}) also enable the manipulation of faces.  DeepFakes can be abused maliciously, such as by creating revenge pornography \cite{BBC, VICE} or striking at politicians \cite{agarwal2019protecting}. This fake multimedia information can easily deceive the human senses, raising concerns of  privacy, social risks, and even national security. The detection of  DeepFakes to  authenticate multimedia content is a crucial need.

DeepFakes generated by imperfect synthesis algorithms will introduce noticeable visual artifacts, inspiring many artifact-based detection methods \cite{yang2019exposing, li2018ictu}. However, with   improved means of generating DeepFakes, these methods have struggled. Hence, it is necessary to capture more fine-grained  differences between real and fake faces, and this is a specialty of deep neural networks (DNNs), whose  impressive classification performance   has spawned many detection models  \cite{zhou2017two, afchar2018mesonet, nguyen2019use, li2018exposing, rossler2019faceforensics++}. Some models directly take the original image pixels as input, and detection by DNNs automatically extracting features still struggles to meet the challenge of more advanced DeepFakes \cite{li2020celeb}. In addition, a trained detector is likely based on the specific features of one dataset, and cannot extrapolate to other datasets, i.e., it lacks the ability to generalize. To more effectively detect DeepFakes, some recent work has meticulously designed DNNs to combine modules or features with positive detection capabilities, such as an attention mechanism \cite{dang2020detection, zhao2021multi, zi2020wilddeepfake}, texture features \cite{liu2020global, sun2020identifying}, audio and visual modalities \cite{chugh2020not}, and frequency spectrum  \cite{zhang2019detecting}. Concomitantly, to drive large and complex DNNs with large-scale datasets requires significant computing resources (e.g., GPUs) and training (e.g., parameter adjustment), which decrease efficiency.

\begin{figure}[h]
  \centering
  \includegraphics[width=\linewidth]{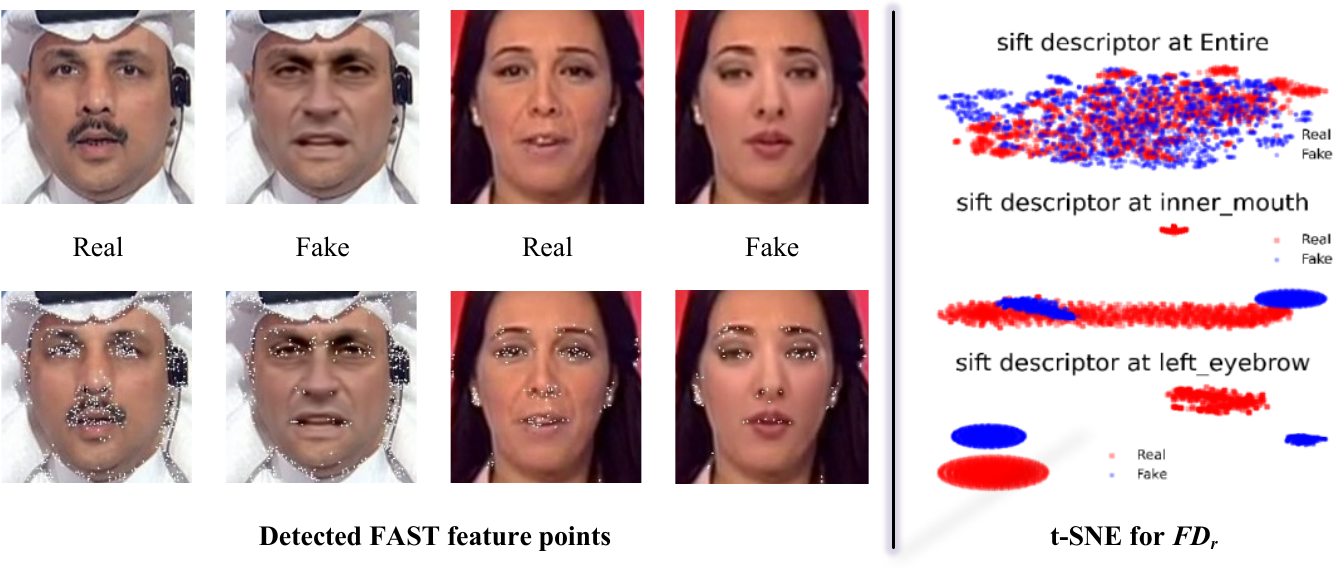}
  \caption{The left side shows two real images and two fake images from  FaceForensics++\_DeepFakes \cite{rossler2019faceforensics++}. The white dots in the corresponding images below are the detected FAST \cite{rosten2006machine} feature points. It can be seen that fake faces have fewer feature points than real faces. T-SNE \cite{van2008visualizing} of $FD_{r}$ in the entire face, inner\_mouth, and left\_eyebrow region is shown on the right. $FD_{r}$ is constructed by a SIFT \cite{lowe2004distinctive} detector-descriptor and comes from   2000 real and 2000 fake images that were randomly selected.}
\label{fig:intro}
\end{figure}

To quickly and effectively detect DeepFakes, we aim to capture  subtle defects of local features. Image local feature detection-description algorithms have been successfully used in various vision applications, such as  multimedia content retrieval, image matching, object detection, video data mining, and augmented reality \cite{mian2010repeatability, kim2012local, marchand2015pose, tareen2018comparative, krig2016interest}. Algorithms to determine   feature points by specifying the properties of images (e.g., edge and corner) are called detectors, and  algorithms that describe   detected points by neighboring pixel regions are referred to as descriptors. Motivated by the advantages of feature point detector-descriptors, such as fast detection and low-dimensional yet discriminative descriptions, we explore information about the number, distribution, and description of feature points in DeepFake faces.

Deep generative models used for face swaps, such as autoencoders \cite{Faceswap}, require the reconstruction of faces from latent features. However, the limited encoding space will cause the replaced face to be slightly blurred \cite{afchar2018mesonet}, which will affect the feature point detector that makes decisions through pixel-level differences. Our experiments on  self-swapped faces (see Section \ref{sec:swapself}) and the widely used DeepFake datasets \cite{li2018ictu, korshunov2018deepfakes, rossler2019faceforensics++, dufour2019contributing, dolhansky2019deepfake, li2020celeb} reveal that the number of feature points in DeepFake faces is generally less than in real ones, especially in the eyes, nose, mouth, and eyebrows. Indeed, due to the concentration of details in these facial regions, the feature points of faces tend to be distributed in them, as shown on the left of Figure \ref{fig:intro}. However, descriptors that record fine-grained features such as gradient information \cite{lowe2004distinctive} or intensity differences \cite{calonder2010brief, rublee2011orb} will also be affected by blur. Each feature point of one face will be described by one descriptor, making it arduous to detect DeepFake through a mass of messy descriptors. Based on the above findings, we propose the Fused Facial Region\_Feature Descriptor (FFR\_FD), a compact vector representation to detect fake faces.

FFR\_FD can be constructed by various feature point detector-descriptors, and our experiments include SIFT \cite{lowe2004distinctive}, SURF \cite{bay2008speeded}, FAST\&BRIEF \cite{rosten2006machine, calonder2010brief}, ORB \cite{rublee2011orb}, and A-KAZE \cite{alcantarilla2011fast}. We divide  feature points   into eight facial regions: entire face, mouth, inner\_mouth, right\_eyebrow, left\_eyebrow,  right\_eye, left\_eye, and nose. All of the feature point descriptors in each facial region are used to construct a vector for that region,  $FD_{r}$, which can be controlled to introduce the quantity information of feature points by not taking the average value during the calculation. We connect $FD_{r}$ for each  facial region  in series to form FFR\_FD, which is consistent with the computational simplicity of   feature detector-descriptors, and   significantly reduces the dimension of features extracted from each face, from an $N$ x  $d$ matrix to an $8$  x $d$ vector, where $N$ is the total number of feature points in an image and $d$ is the descriptor's dimension. To subdivide facial regions with $FD_{r}$  capitalizes on the  lack of feature points of DeepFakes in these regions, and is more discriminative than to directly use all descriptors in the entire face, as shown on the right of Figure \ref{fig:intro}.

We used FFR\_FD as input to train the random forest classifier. Extensive experiments were conducted on six DeepFake benchmark datasets. Results demonstrate the effectiveness of our lightweight model for DeepFake detection. Specifically, the frame-level AUC scores of the challenging DFD \cite{dufour2019contributing}, DFDC \cite{dolhansky2019deepfake}, and CelebDF\_V2 \cite{li2020celeb} reached 85.2, 88.3, and 82.2, respectively,  outperforming other DNN-based state of the art (SOTA) detection methods. In addition, our method has considerable generalizability, which is useful given that current DeepFake faces generally lack feature points. Our main contributions are as follows:

$\bullet$ Our empirical studies reveal that the use of deep generative models to swap faces will reduce the detected feature points. The detection and statistical results of many large-scale datasets show that  current DeepFake faces have fewer feature points than real faces, especially in detailed  regions. This substantial difference is evident even on higher-quality datasets.

$\bullet$ We propose FFR\_FD as an informative vector to represent the facial feature description, constructed by the feature point descriptors from the subdivided facial regions. FFR\_FD has   computational simplicity and can be combined with various feature point detector-descriptors.

$\bullet$ Random forest trained with FFR\_FD can realize state-of-the-art detection. Compared with popular DNN-based detection models, our method is more effective, efficient, and generalizable.

\section{Related work}
We introduce  DeepFake generation and detection methods, and describe the DeepFake datasets and typical feature point detector-descriptors used in our experiments.

\subsection{DeepFake generation and datasets}
DeepFake, as a general term for face swapping, refers to fake images or videos synthesized by algorithms, such as GANs \cite{goodfellow2014generative, karras2019style, nirkin2019fsgan, karras2020analyzing}, auto-encoders \cite{kingma2013auto, tewari2017mofa, Faceswap}, and 3D-models \cite{thies2016face2face, thies2019deferred}. The rapid progress of these deep generative models presents significant challenges to the quick and effective detection of realistic DeepFakes. Many DeepFake video benchmark datasets have been released to promote the research of DeepFake detection. Current DeepFake datasets basically comprise two generations \cite{li2020celeb, tolosana2020deepfakes}. The first includes UADFV \cite{li2018ictu}, DeepfakeTIMIT \cite{korshunov2018deepfakes}, and FF++\_DeepFakes \cite{rossler2019faceforensics++}, using FakeApp \cite{FakeAPP}, Faceswap-Gan \cite{faceswap-GAN}, and Faceswap  \cite{Faceswap} to generate fake videos. The second generation, including DFD \cite{dufour2019contributing}, DFDC \cite{dolhansky2019deepfake}, and Celeb-DF(V2) \cite{li2020celeb},   improves  visual quality   through augmented synthesis algorithms. The number of videos has increased, along with their diversity, such as age, skin color, and lighting environment. Therefore, we cannot ignore the cost and speed of a detection model in addition to its performance.

\subsection{DeepFake detection}
Early work identified physical behavior patterns, such as inconsistent head poses \cite{yang2019exposing},   unnatural eye blinking \cite{li2018ictu}, and  correlations between facial expressions and head movements \cite{agarwal2019protecting}. 
However, these artifacts were fixed in   second-generation DeepFake datasets, resulting in limited detection performance. Recent work has also exposed DeepFakes based on biological signals \cite{ciftci2020fakecatcher}.

Detection methods based on deep neural networks (DNNs) have become mainstream. For example, a two-stream CNN was used  \cite{zhou2017two}, Meso-4   focused on the mesoscopic properties of images \cite{afchar2018mesonet}, a capsule structure  based on VGG19 \cite{nguyen2019use} was used, ResNet was used to capture faces warping artifacts \cite{li2018exposing}, and classic Xception \cite{rossler2019faceforensics++, marra2018detection} was used to detect fake faces. Because videos have temporal features, some researchers have combined CNNs with RNNs for classification \cite{amerini2020exploiting, montserrat2020deepfakes}. With their powerful feature extraction capabilities, DNN-based methods have achieved some success, but they still have limitations against advanced DeepFakes \cite{li2020celeb}. Learning-based methods have been further studied bo address this issue. For example, FakeSpotter \cite{wang2019fakespotter} monitors neuron behavior \cite{xie2019deephunter} to detect fake faces. More recently,   researchers have  combined useful modules or important features. Dang et al. \cite{dang2020detection} utilized an attention mechanism to improve   detection ability. Similarly, a vision transformer was used for detection \cite{wodajo2021deepfake}. Gram-Net \cite{liu2020global} and InTeLe \cite{sun2020identifying} explore the texture information of images to improve robustness. A method combining an attention mechanism and texture features was   proposed \cite{zhao2021multi}. Instead of designing large, complex neural networks, we efficiently extract features for effective DeepFake detection.

To improve generalization ability, Cozzolino et al. \cite{cozzolino2018forensictransfer} proposed to learn an embedding based on an autoencoder. Wang et al. \cite{wang2020cnn} trained   ResNet \cite{he2016deep} with a multi-class ProGAN dataset and showed that appropriate preprocessing and postprocessing could improve generalization. Face X-ray \cite{li2020face} observes the blending boundaries between   faces and the background to detect swapped faces; its framework adopts   HRNet \cite{sun2019deep, sun2019high}. The usampling strategies of deep generative models   introduce artifacts in the frequency domain \cite{durall2020watch, frank2020leveraging}, inspiring many spectrum-based detection methods \cite{zhang2019detecting, durall2019unmasking}. However,   detection based  only  on the frequency spectrum leads to unsatisfactory performance and generalization. Frequency-domain artifacts can be reduced by training with  spectrum regularization \cite{durall2020watch},   focal frequency loss \cite{jiang2020focal}, or a spectrum discriminator \cite{jung2020spectral}. FakePolisher \cite{huang2020fakepolisher} performs shallow reconstruction and can   reduce   artifact patterns. This calls for the discovery of the more fine-grained feature defects of DeepFakes to provide effective DeepFake detection.

\subsection{Feature point detector-descriptors}
The detection and description of local features of images is a fundamental problem in   computer vision \cite{wang2018facial, karami2017image}. Feature points (also known as interest points or keypoints) are widely used in image matching, retrieval, and recognition tasks. 

\noindent\textbf{SIFT} The scale invariant feature transform (SIFT) is a well-known feature detection and description algorithm proposed by Lowe \cite{lowe2004distinctive}. It has four steps: scale-space extrema detection, i.e., the use of difference of Gaussian (DOG) to search for maxima at various scales of the image; keypoint localization, i.e., the location of potential feature points, where the Hessian matrix is used to remove low-contrast points; orientation assignment, i.e., the construction of an orientation histogram based on the gradient information around the feature point, and its use  to assign orientation for the feature point; and the keypoint descriptor, i.e., use of the gradient magnitude and orientation to construct a feature vector for feature points. This method extracts a 16 x 16 neighbor region in each, and subdivides it into 4 x 4 subblocks with eight orientation bins, resulting in a descriptor of 4 x 4 x 8 = 128 dimensions.

\noindent\textbf{SURF} The feature-detection  of speeded up robust features (SURF),  proposed by Bay et al. \cite{bay2008speeded}, is faster than that of SIFT. Due to the use of integral images, the boxed filter approximates   DOG. SURF uses the determinant of the Hessian matrix to find the interest points at blob-type structures. 2D Haar wavelet responses are used for orientation assignment and feature description. The neighborhood around each feature point is divided into 4 x 4 subregions, and every wavelet response of a subregion has four values, so the descriptor of each feature point is a vector of dimension  4 x 4 x 4=64.

\noindent\textbf{FAST\&BRIEF}  Features from Accelerated Segment Test (FAST) \cite{rosten2006machine} is a high-speed feature detector developed for real-time applications. Corner point extraction is performed by comparing the intensity thresholds between the point and pixels on a circular ring around it. A decision tree based on the ID3 algorithm is trained as the corner detector. Non-maximum suppression is used to solve the problem of adjacent corner points, which makes the FAST detector more efficient. FAST is only a feature point detector and does not involve feature description.  Binary Robust Independent Elementary Features (BRIEF) \cite{calonder2010brief} is a feature point descriptor that uses simple binary strings. It performs Gaussian smoothing for the patch partition of detected feature points, and   randomly selects pairs of points for intensity difference tests to obtain descriptors. If the number of pairs of points is 256, then the dimension of the descriptor is 32   bytes for each feature point, which is much smaller than that of   SIFT or SURF.  BRIEF performs similarly to SIFT in most scenarios, but its low complexity makes the construction of descriptors faster. Combining the FAST detector and  BRIEF descriptor has the advantages of fast speed and low computational resource requirements.

\noindent\textbf{ORB}  Introduced by Rublee et al. \cite{rublee2011orb}, Oriented FAST and Rotated BRIEF (ORB) is mainly used to solve the problem that the FAST detector does not compute orientation and the BRIEF descriptor lacks rotational invariance. ORB applies the Harris corner measure to pick the Top $N$ points from FAST corners to evaluate cornerness and determine the local orientation by the intensity centroid \cite{rublee2011orb}. ORB extracts the BRIEF descriptor according to the main orientation of the feature point. It adopts a greedy algorithm to find random pairwise pixel patches with low correlation. In this study, the authors selected 256 pairwise pixel patches for feature description, so it generates a 32-dimensional descriptor. Note that we use the feature point information for DeepFake detection. The extracted faces have been cropped and aligned, so the improved properties of ORB have little impact on  detection, but we still use it for comparison.

\noindent\textbf{A-KAZE} Introduced by Alcantarilla et al. \cite{alcantarilla2011fast}, the A-KAZE feature detector and descriptor effectively constructs the nonlinear scale space for feature detection through FED \cite{weickert2016cyclic, grewenig2010box}. The descriptor   is based on M-LDB, which is also efficient at exploring gradient information from the nonlinear scale space.

The features extracted by the detector have different property requirements, depending on the vision task~\cite{tareen2018comparative}. We are concerned about the effectiveness and efficiency of using it for DeepFake detection.

\section{Feature points of DeepFakes and real faces} \label{sec:FP_num}

In this section, we show the number and distribution of feature points between DeepFake faces and real ones in detail.

\subsection{Swapping faces reduces the number of feature points} \label{sec:swapself}

\begin{figure}[htb]
  \centering
  \includegraphics[width=1\linewidth]{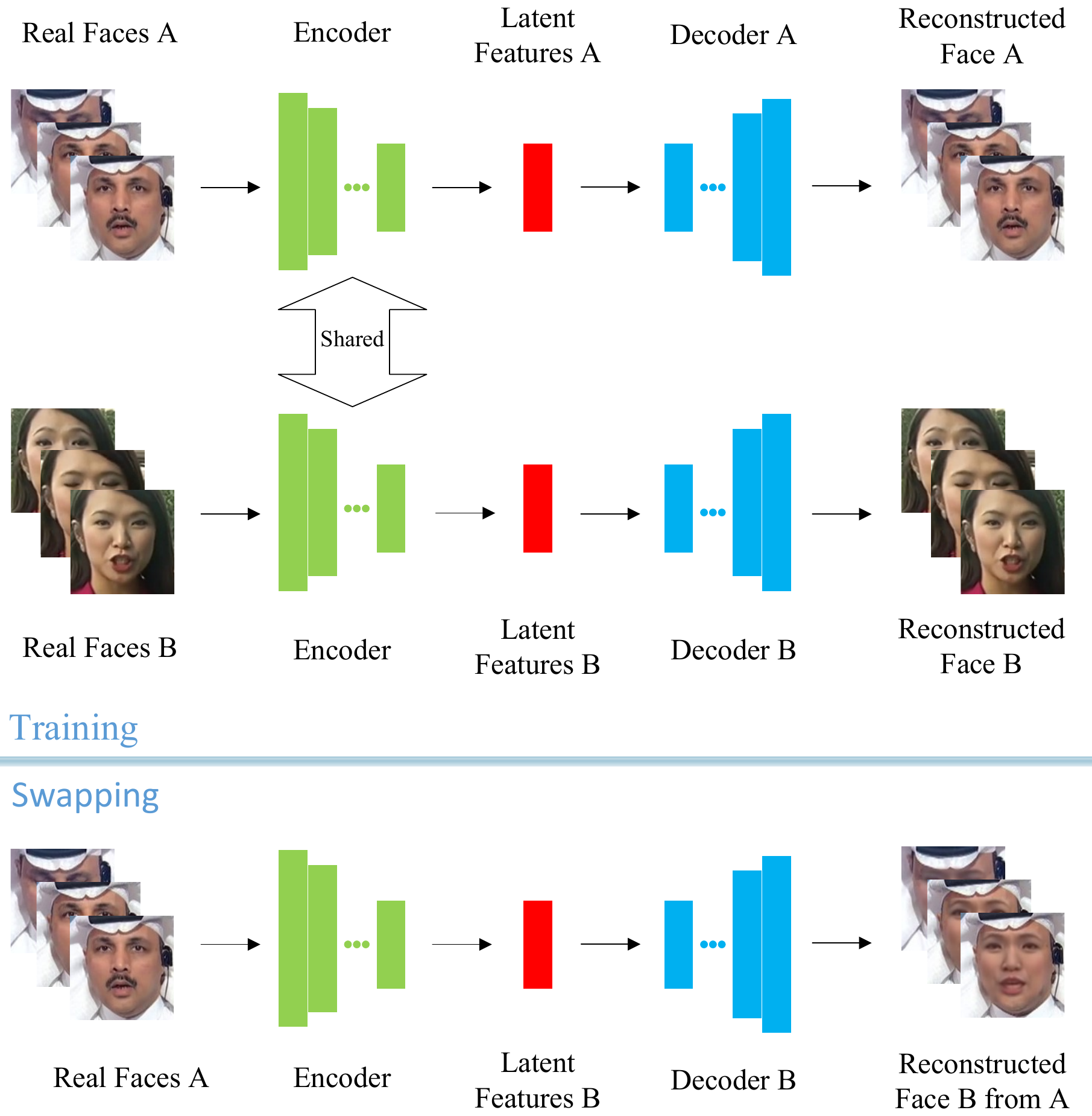}
  \caption{Procedures to create a DeepFake based on an autoencoder-decoder}
\label{fig:AE}
\end{figure}

Will deep generative models cause a change in the number of facial feature points? To this end, we utilize the \emph{Original Faceswap Model}  \cite{Faceswap} for face swapping. The model is based on the autoencoder-decoder: the shared encoder is trained for two different faces, \emph{A} and \emph{B}, to extract the latent feature vectors, from which the corresponding faces  are reconstructed by their respective decoders. The compressed latent vectors can be understood as general face information such as the expression and background, and the decoders are used to restore specific facial features, such as nose details, from latent vectors. After training, the latent vectors extracted from \emph{A}  by the shared encoder are transmitted to the decoder of  \emph{B}, such that \emph{A} is replaced by   \emph{B}, as shown in Figure \ref{fig:AE}.

\begin{figure}[htb]
  \centering
  \includegraphics[width=\linewidth]{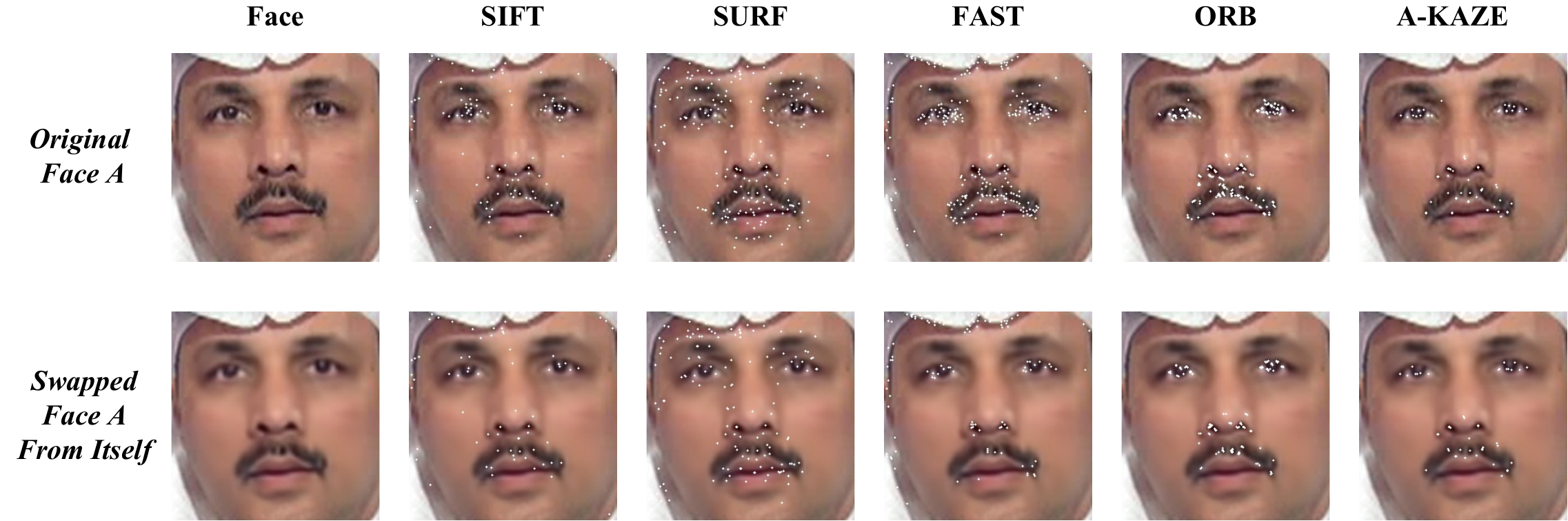}
  \caption{Feature points detected in  original   and   swapped faces.  SIFT, SURF, FAST, ORB, and A-KAZE feature-point detectors are used in experiments. Comparing the corresponding feature point detectors of original and swapped faces, the DeepFake lacks feature points, especially in facial regions such as the eyes, nose, and mouth.}
\label{fig:fp_swap_self}
\end{figure}

Because the feature points of diverse facial appearances are naturally different, we use   one face as both \emph{A} and \emph{B}, i.e., we swap a face with itself. A real face video containing 396 frames from FF++\_YouTube \cite{rossler2019faceforensics++} was used for experiments. After sufficient training of the model, we swapped the face with itself and detected the SIFT, SURF, FAST, ORB, and A-KAZE feature points. The feature points of the original face and the corresponding fake face are shown in Figure \ref{fig:fp_swap_self}. Obviously, the fake face has fewer feature points   than  the   real face with the same feature point detector. Examining the distribution of feature points, this phenomenon is more noticeable in facial regions with detailed information, such as the nose, mouth, eyes, and eyebrows. It can also be seen from Figure \ref{fig:fp_swap_self} that while the swapped face is similar to the original, many refined details for determining feature points are lost.

To verify the observation results, we counted the numbers of feature points on the original and swapped faces. We  summed the feature points in the face images extracted from all of the frames  and divided by the number of faces to obtain the average number of feature points of each face. Based on the discrepancies of the feature points distribution, we further used Dlib \cite{king2009dlib} to subdivide the face into seven regions: mouth, inner\_mouth, right\_eyebrow, left\_eyebrow, right\_eye, left\_eye, and nose; then we calculated the average number of feature points, as shown in Table \ref{tab:FP_NUM_swapself}. These   results   again show that the Faceswap model will reduce the feature points of faces, especially in the   subdivided facial regions. Compared to other detectors, the number of feature points of fake faces is reduced more significantly on   FAST and ORB (based on modified FAST). Both visualizations and statistical results inspire us to explore the feature points on detailed facial regions for detecting forged faces.

\begin{table}[!htb]
  \caption{Average number of feature points for eight facial regions. Results come from   original faces and   corresponding self-swapped faces. Real and fake faces are extracted from the corresponding 369 frames.}
  \label{tab:FP_NUM_swapself}
  \centering
  \resizebox{\linewidth}{!}{%
  \begin{tabular}{c|cc|cc|cc|cc|cc} 
\hline
\textit{FP Detector~~}→           & \multicolumn{2}{c|}{\textbf{SIFT}} & \multicolumn{2}{c|}{\textbf{SURF}} & \multicolumn{2}{c|}{\textbf{FAST}} & \multicolumn{2}{c|}{\textbf{ORB}} & \multicolumn{2}{c}{\textbf{A-KAZE}}  \\
\textit{Region~}↓ \textit{Face }→ & ori   & swap                       & ori   & swap                       & ori   & swap                       & ori   & swap                      & ori  & swap                          \\ 
\hline\hline
entire face                       & 126.7 & 91.6                       & 182.6 & 141.1                      & 146.1 & 53.7                       & 237.2 & 108.6                     & 67.5 & 50.7                          \\
mouth                             & 23.9  & 10.1                       & 43.3  & 29.1                       & 46.4  & 8.7                        & 70.2  & 17.9                      & 23.8 & 14.5                          \\
inner\_mouth                      & 23.9  & 3.6                        & 18.8  & 13.0                       & 21.8  & 2.9                        & 32.7  & 17.9                      & 9.4  & 3.3                           \\
right\_eyebrow                    & 4.2   & 3.0                        & 18.8  & 9.8                        & 3.5   & 0.4                        & 1.2   & 0.1                       & 0.6  & 0.3                           \\
left\_eyebrow                     & 4.2   & 2.2                        & 13.2  & 8.0                        & 7.6   & 2.3                        & 6.1   & 3.3                       & 1.6  & 1.1                           \\
right\_eye                        & 10.4  & 4.4                        & 8.5   & 6.5                        & 21.1  & 7.8                        & 37.5  & 15.4                      & 8.8  & 7.4                           \\
left\_eye                         & 11.0  & 6.7                        & 8.8   & 7.4                        & 20.0  & 10.4                       & 39.8  & 21.0                      & 8.7  & 7.8                           \\
nose                              & 11.0  & 10.4                       & 24.0  & 18.8                       & 23.1  & 12.3                       & 55.3  & 33.7                      & 8.7  & 11.6                          \\
\hline
\end{tabular}%
}
\end{table}

\begin{table*}[!htb]
   \caption{Average number of feature points for faces from  DeepFake datasets. All images from the training set   are used for statistics. Based on five feature detector-descriptors,   current DeepFakes struggle with providing enough feature points across facial regions. DT (HQ) and DT (LQ) denote the HQ and LQ versions, respectively, of DeepfakeTIMIT \cite{korshunov2018deepfakes}. FF++\_DF (RAW) and FF++\_DF (LQ) denote the uncompressed and \textit{c40} high-compressed versions, respectively, of FF++\_DeepFakes \cite{rossler2019faceforensics++}.}
   \label{tab:FP_NUM_datasets}
   \centering
   \resizebox{\textwidth}{!}{%
   \begin{tabular}{c|c|cc|cc|cc|cc|cc|c|c|cc|cc|cc|cc|cc} 
\hline
{\cellcolor[rgb]{0.753,0.753,0.753}}                                                                         & {\cellcolor[rgb]{0.753,0.753,0.753}}\textit{Detector~}→ & \multicolumn{2}{c|}{\textbf{SIFT}} & \multicolumn{2}{c|}{\textbf{SURF}} & \multicolumn{2}{c|}{\textbf{FAST}} & \multicolumn{2}{c|}{\textbf{ORB}} & \multicolumn{2}{c|}{\textbf{A-KAZE}} & {\cellcolor[rgb]{0.753,0.753,0.753}}                                                                         & {\cellcolor[rgb]{0.753,0.753,0.753}}\textit{Detector} → & \multicolumn{2}{c|}{\textbf{SIFT}} & \multicolumn{2}{c|}{\textbf{SURF}} & \multicolumn{2}{c|}{\textbf{FAST}} & \multicolumn{2}{c|}{\textbf{ORB}} & \multicolumn{2}{c}{\textbf{A-KAZE}}  \\
\multirow{-2}{*}{{\cellcolor[rgb]{0.753,0.753,0.753}}\textit{Dataset}~ ↓}                                    & {\cellcolor[rgb]{0.753,0.753,0.753}}\textit{Region} ↓   & real  & fake                       & real  & fake                       & real  & fake                       & real  & fake                      & real & fake                          & \multirow{-2}{*}{{\cellcolor[rgb]{0.753,0.753,0.753}}\textit{Dataset} ↓}                                     & {\cellcolor[rgb]{0.753,0.753,0.753}}\textit{Region }↓   & real  & fake                       & real  & fake                       & real  & fake                       & real  & fake                      & real & fake                          \\ 
\hline\hline
\multirow{8}{*}{\textbf{\textit{DT(HQ)}}}                                                                    & entire Face                                             & 77.9  & 68.9                       & 173.9 & 159.7                      & 99.2  & 65.7                       & 107.1 & 88.1                      & 33.9 & 34.3                          & \multirow{8}{*}{\begin{tabular}[c]{@{}c@{}}\textbf{\textit{FF++\_DF}}\\\textbf{\textit{(LQ)}}\end{tabular}} & entire Face                                             & 94.6  & 81.1                       & 181.7 & 169.3                      & 123.7 & 91.6                       & 161.5 & 98.8                      & 48.4 & 37.5                          \\
                                                                                                             & mouth                                                   & 8.0   & 5.2                        & 22.5  & 19.5                       & 13.6  & 6.6                        & 15.8  & 9.9                       & 5.7  & 5.0                           &                                                                                                              & mouth                                                   & 9.0   & 4.9                        & 19.3  & 14.8                       & 20.3  & 8.6                        & 31.3  & 13.0                      & 8.9  & 5.3                           \\
                                                                                                             & inner\_mouth                                            & 4.4   & 2.8                        & 7.8   & 6.6                        & 8.5   & 4.0                        & 11.4  & 7.2                       & 3.6  & 3.0                           &                                                                                                              & inner\_mouth                                            & 9.0   & 2.7                        & 7.2   & 5.2                        & 12.7  & 5.3                        & 22.6  & 9.2                       & 5.8  & 3.4                           \\
                                                                                                             & right\_eyebrow                                          & 1.4   & 1.3                        & 5.9   & 5.2                        & 2.5   & 1.5                        & 1.0   & 0.7                       & 0.9  & 0.9                           &                                                                                                              & right\_eyebrow                                          & 2.9   & 2.4                        & 7.8   & 6.9                        & 6.6   & 5.4                        & 5.9   & 4.3                       & 3.0  & 2.5                           \\
                                                                                                             & left\_eyebrow                                           & 1.5   & 1.4                        & 6.3   & 5.7                        & 3.0   & 2.1                        & 1.0   & 1.1                       & 0.8  & 1.2                           &                                                                                                              & left\_eyebrow                                           & 2.9   & 2.4                        & 7.9   & 6.7                        & 6.4   & 5.2                        & 5.5   & 4.4                       & 2.9  & 2.4                           \\
                                                                                                             & right\_eye                                              & 4.1   & 3.2                        & 5.5   & 4.9                        & 9.9   & 6.4                        & 19.6  & 14.5                      & 0.8  & 3.8                           &                                                                                                              & right\_eye                                              & 5.5   & 2.9                        & 5.2   & 4.2                        & 11.0  & 6.1                        & 24.6  & 13.0                      & 5.4  & 3.9                           \\
                                                                                                             & left\_eye                                               & 4.4   & 3.2                        & 5.3   & 5.0                        & 10.5  & 6.6                        & 21.1  & 16.0                      & 4.2  & 4.2                           &                                                                                                              & left\_eye                                               & 5.4   & 2.9                        & 5.3   & 4.3                        & 10.9  & 6.0                        & 25.0  & 13.2                      & 5.5  & 4.0                           \\
                                                                                                             & nose                                                    & 8.0   & 7.7                        & 15.8  & 15.6                       & 13.4  & 9.6                        & 28.7  & 29.8                      & 11.4 & 12.4                          &                                                                                                              & nose                                                    & 7.6   & 5.6                        & 14.8  & 12.3                       & 10.5  & 5.5                        & 26.8  & 15.7                      & 9.3  & 7.2                           \\ 
\hline
\multirow{8}{*}{\textbf{\textit{DT(LQ)}}}                                                                    & entire Face                                             & 77.9  & 57.9                       & 173.9 & 130.2                      & 99.2  & 38.6                       & 107.1 & 45.2                      & 33.9 & 21.1                          & \multirow{8}{*}{\textbf{\textit{DFD}}}                                                                       & entire Face                                             & 118.7 & 100.2                      & 195.9 & 183.5                      & 193.8 & 131.4                      & 208.2 & 160.0                     & 52.7 & 50.0                          \\
                                                                                                             & mouth                                                   & 8.0   & 3.5                        & 22.5  & 15.2                       & 13.6  & 2.1                        & 15.8  & 3.4                       & 5.7  & 2.3                           &                                                                                                              & mouth                                                   & 15.1  & 7.8                        & 26.3  & 23.0                       & 32.1  & 13.4                       & 46.9  & 25.4                      & 11.6 & 9.7                           \\
                                                                                                             & inner\_mouth                                            & 4.4   & 1.9                        & 7.8   & 4.6                        & 8.5   & 1.6                        & 11.4  & 2.8                       & 3.6  & 1.6                           &                                                                                                              & inner\_mouth                                            & 9.9   & 4.4                        & 10.7  & 9.5                        & 20.7  & 13.4                       & 35.9  & 17.7                      & 7.9  & 6.3                           \\
                                                                                                             & right\_eyebrow                                          & 1.4   & 0.8                        & 5.9   & 3.0                        & 2.5   & 0.2                        & 1.0   & 0.1                       & 0.9  & 0.1                           &                                                                                                              & right\_eyebrow                                          & 2.6   & 1.9                        & 6.8   & 6.0                        & 7.7   & 4.5                        & 4.7   & 3.5                       & 1.7  & 1.5                           \\
                                                                                                             & left\_eyebrow                                           & 1.5   & 0.8                        & 6.3   & 3.6                        & 3.0   & 0.3                        & 1.0   & 0.2                       & 0.8  & 0.3                           &                                                                                                              & left\_eyebrow                                           & 2.5   & 2.0                        & 7.6   & 6.3                        & 8.3   & 4.7                        & 4.5   & 3.2                       & 1.5  & 1.3                           \\
                                                                                                             & right\_eye                                              & 4.1   & 2.2                        & 5.5   & 4.5                        & 9.9   & 3.1                        & 19.6  & 7.2                       & 0.8  & 2.4                           &                                                                                                              & right\_eye                                              & 6.1   & 4.3                        & 5.4   & 5.0                        & 13.7  & 8.4                        & 24.2  & 18.0                      & 4.6  & 4.6                           \\
                                                                                                             & left\_eye                                               & 4.4   & 2.2                        & 5.3   & 4.6                        & 10.5  & 3.5                        & 21.1  & 8.2                       & 4.2  & 2.6                           &                                                                                                              & left\_eye                                               & 6.1   & 4.5                        & 5.6   & 5.2                        & 14.3  & 8.9                        & 25.0  & 19.3                      & 4.7  & 4.9                           \\
                                                                                                             & nose                                                    & 8.0   & 5.9                        & 15.8  & 11.7                       & 13.4  & 3.4                        & 28.7  & 15.1                      & 11.4 & 9.1                           &                                                                                                              & nose                                                    & 10.8  & 9.1                        & 19.6  & 17.0                       & 14.3  & 15.1                       & 46.5  & 38.7                      & 14.0 & 13.6                          \\ 
\hline
\multirow{8}{*}{\textbf{\textit{UADFV}}}                                                                     & entire Face                                             & 96.4  & 74.8                       & 186.1 & 160.6                      & 108.2 & 116.7                      & 148.3 & 71.9                      & 40.8 & 22.3                          & \multirow{8}{*}{\textbf{\textit{DFDC}}}                                                                      & entire Face                                             & 83.0  & 70.0                       & 152.8 & 131.7                      & 99.4  & 78.8                       & 134.1 & 94.9                      & 41.5 & 28.2                          \\
                                                                                                             & mouth                                                   & 9.3   & 3.0                        & 20.0  & 11.8                       & 16.2  & 14.1                       & 27.2  & 9.3                       & 8.1  & 2.5                           &                                                                                                              & mouth                                                   & 9.0   & 7.0                        & 18.7  & 16.5                       & 16.1  & 12.4                       & 25.8  & 18.8                      & 7.7  & 5.2                           \\
                                                                                                             & inner\_mouth                                            & 5.6   & 2.1                        & 7.1   & 4.6                        & 9.7   & 8.4                        & 18.8  & 6.4                       & 5.3  & 1.8                           &                                                                                                              & inner\_mouth                                            & 6.1   & 4.8                        & 8.6   & 7.7                        & 11.6  & 9.4                        & 20.8  & 15.7                      & 5.4  & 3.9                           \\
                                                                                                             & right\_eyebrow                                          & 2.4   & 2.7                        & 8.6   & 4.6                        & 4.4   & 9.8                        & 3.4   & 1.7                       & 2.0  & 1.5                           &                                                                                                              & right\_eyebrow                                          & 2.1   & 1.7                        & 7.1   & 6.4                        & 3.3   & 2.9                        & 3.2   & 2.3                       & 1.4  & 0.9                           \\
                                                                                                             & left\_eyebrow                                           & 2.3   & 2.4                        & 7.9   & 8.2                        & 4.7   & 9.9                        & 3.8   & 2.3                       & 2.0  & 1.4                           &                                                                                                              & left\_eyebrow                                           & 2.2   & 1.6                        & 8.3   & 6.4                        & 4.2   & 2.4                        & 3.3   & 1.5                       & 2.1  & 0.6                           \\
                                                                                                             & right\_eye                                              & 6.1   & 3.3                        & 4.8   & 4.4                        & 11.4  & 10.5                       & 23.8  & 10.9                      & 4.7  & 2.9                           &                                                                                                              & right\_eye                                              & 4.9   & 4.2                        & 5.4   & 5.0                        & 6.8   & 6.1                        & 13.4  & 11.7                      & 3.4  & 2.7                           \\
                                                                                                             & left\_eye                                               & 5.8   & 3.0                        & 4.8   & 4.6                        & 11.2  & 10.3                       & 24.4  & 11.9                      & 4.6  & 2.7                           &                                                                                                              & left\_eye                                               & 4.9   & 4.1                        & 5.9   & 5.3                        & 7.2   & 6.0                        & 14.9  & 11.7                      & 3.4  & 2.6                           \\
                                                                                                             & nose                                                    & 10.1  & 6.0                        & 19.5  & 18.9                       & 14.8  & 15.5                       & 36.4  & 14.9                      & 11.8 & 6.0                           &                                                                                                              & nose                                                    & 9.5   & 7.8                        & 17.8  & 16.7                       & 14.0  & 12.3                       & 34.6  & 27.3                      & 10.7 & 9.0                           \\ 
\hline
\multirow{8}{*}{\begin{tabular}[c]{@{}c@{}}\textbf{\textit{FF++\_DF}}\\\textbf{\textit{(RAW)}}\end{tabular}} & entire Face                                             & 122.4 & 100.3                      & 209.4 & 190.5                      & 158.8 & 98.2                       & 215.1 & 128.8                     & 57.5 & 44.5                          & \multirow{8}{*}{\textbf{\textit{CelebDF\_V2}}}                                                               & entire Face                                             & 58.0  & 46.2                       & 140.1 & 121.3                      & 54.8  & 31.7                       & 87.2  & 46.8                      & 34.3 & 22.7                          \\
                                                                                                             & mouth                                                   & 12.7  & 5.3                        & 22.6  & 16.6                       & 26.2  & 6.4                        & 42.6  & 15.1                      & 10.9 & 6.3                           &                                                                                                              & mouth                                                   & 7.2   & 3.6                        & 18.8  & 14.9                       & 10.3  & 4.0                        & 18.3  & 6.9                       & 6.8  & 3.4                           \\
                                                                                                             & inner\_mouth                                            & 8.0   & 3.0                        & 8.3   & 5.9                        & 16.4  & 4.4                        & 31.7  & 11.0                      & 7.3  & 4.1                           &                                                                                                              & inner\_mouth                                            & 7.2   & 1.9                        & 7.5   & 5.1                        & 7.2   & 2.6                        & 14.1  & 4.8                       & 6.8  & 2.4                           \\
                                                                                                             & right\_eyebrow                                          & 3.1   & 2.4                        & 8.7   & 7.3                        & 7.8   & 4.6                        & 6.8   & 4.1                       & 3.4  & 2.6                           &                                                                                                              & right\_eyebrow                                          & 1.5   & 0.9                        & 5.6   & 3.9                        & 2.0   & 0.6                        & 1.8   & 0.4                       & 1.4  & 0.7                           \\
                                                                                                             & left\_eyebrow                                           & 3.1   & 2.4                        & 9.0   & 7.2                        & 7.7   & 4.5                        & 6.4   & 4.1                       & 3.3  & 2.5                           &                                                                                                              & left\_eyebrow                                           & 1.5   & 0.9                        & 5.8   & 3.9                        & 2.0   & 0.6                        & 1.7   & 0.4                       & 1.4  & 0.6                           \\
                                                                                                             & right\_eye                                              & 8.2   & 4.9                        & 6.3   & 5.6                        & 15.0  & 8.0                        & 34.7  & 20.9                      & 6.8  & 5.4                           &                                                                                                              & right\_eye                                              & 3.3   & 2.2                        & 4.4   & 3.9                        & 5.8   & 3.4                        & 12.2  & 6.6                       & 3.6  & 2.6                           \\
                                                                                                             & left\_eye                                               & 8.1   & 4.8                        & 6.5   & 5.6                        & 14.9  & 7.8                        & 35.1  & 21.2                      & 7.0  & 5.5                           &                                                                                                              & left\_eye                                               & 3.3   & 2.3                        & 4.5   & 4.1                        & 5.7   & 3.5                        & 12.6  & 7.2                       & 3.7  & 2.8                           \\
                                                                                                             & nose                                                    & 10.7  & 7.5                        & 18.9  & 15.3                       & 16.6  & 7.1                        & 39.8  & 24.0                      & 11.3 & 9.1                           &                                                                                                              & nose                                                    & 7.1   & 5.9                        & 16.6  & 14.6                       & 7.9   & 4.6                        & 23.0  & \multicolumn{1}{c}{14.5}  & 9.6  & 7.6                           \\
\hline
\end{tabular}%
}
\end{table*}

\subsection{Average number of feature points on DeepFake datasets}
Does this defect exist in a DeepFake meticulously crafted with various advanced generation and optimization technologies? We further observed the distribution of feature points on faces from DeepFake datasets, and similarly  divided the facial region to calculate the average number of feature points of real and fake faces, as shown in Table \ref{tab:FP_NUM_datasets}. The results show that both the first- and second-generation DeepFake datasets  lack sufficient feature points. This  motivates us to explore the use of feature point information for detection.

\section{Proposed FFR\_FD for DeepFake detection}

\begin{algorithm}[h!]
\label{FFR_FD_AL}
\renewcommand{\thealgorithm}{4}
\caption{The constrcution of FFR\_FD} 
\hspace*{0.02in} {\bf Input:} 
Input face image $I$\\
\hspace*{0.02in} {\bf Output:} 
$FFR\_FD$ (or $FFR\_FD\_ave$)
\begin{algorithmic}[]
\State 
Set $DET$ to the specified fetaure-point detecor\\
Set $DES$ to the specified feature-point descriptor\\
Set $ROI$ with [mouth, inner-mouth, right-eyebrow, left-eyebrow, right-eye, left-eye, nose]\\
$KP$ = $DET$($I$)
\For{$r$ \textbf{in} $ROI$}
\State $FR[r] = facial\_landmarks(r)$
\EndFor
\For{$r$ \textbf{in} $FR$}
\State $N[r]=0$
\State $FD[r]=0$
\For{$p \in KP$}
\If{$p \in r$}
\State $FD[r]= FD[r] + $Des$(p)$
\State $N[r]=N[r]+1$
\Else
\State \textbf{continue}
\EndIf
\EndFor
\State $FD[r]\_ave= FD[r] / N[r]$
\EndFor
\For{$r \in FR$}
\State $FFR\_FD$ = concatenate($FD[r]$), 
\State $FFR\_FD\_ave$ = concatenate($FD[r]\_ave$)
\EndFor
\State \Return $FFR\_FD$ (or $FFR\_FD\_ave$)
\end{algorithmic}
\end{algorithm}

It is infeasible to directly judge authenticity  by the number of feature points because this fluctuates according to factors such as facial appearance or video quality (compare FF++\_DF (RAW) and FF++\_DF (LQ) in Table \ref{tab:FP_NUM_datasets}). Hence, it is necessary to explore the combination of feature point descriptors for detection, which is  nontrivial if there are many feature points. If $N$ feature points are detected in a face image, and the descriptor of each feature point is $d$-dimensional (depending on the descriptor, as shown in Table \ref{tab:des}), then one-shot will produce ($N$ x $d$)-dimensional features. In current large-scale DeepFake datasets, to directly train the classifier  with descriptors or BOWs \cite{zhang2017automated} (constructed from descriptors by  k-means)   consumes much time  and is prone  to overfitting. Here, we propose FFR\_FD for fast,  effective   DeepFake detection.

\subsection{FFR\_FD} \label{sec:FFR_DD}

\begin{figure*}[h]
  \centering
  \includegraphics[width=\linewidth]{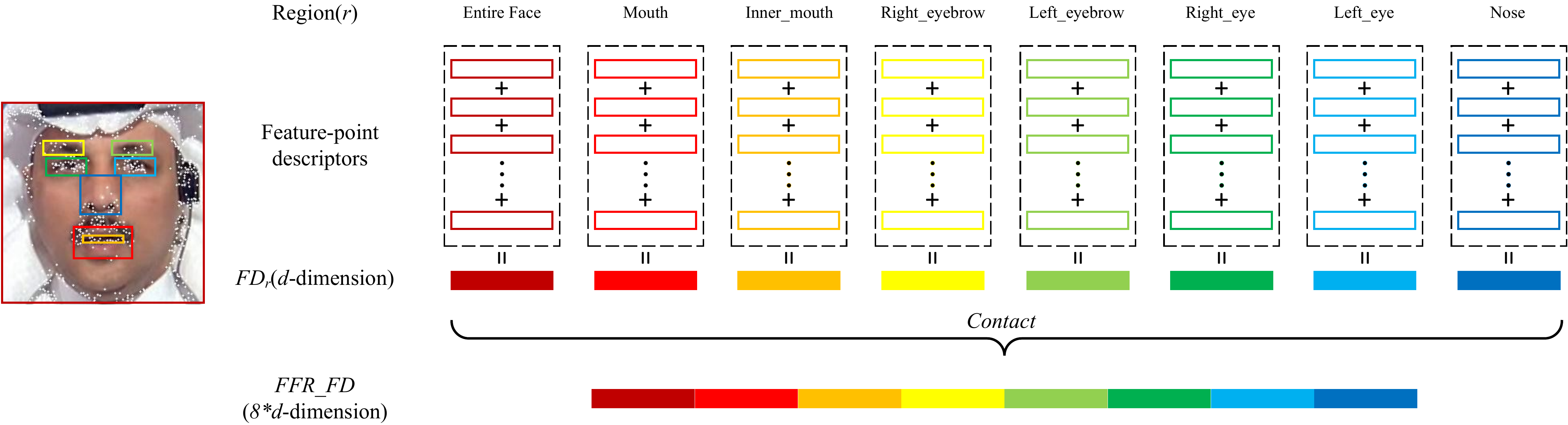}
  \caption{Construction of FFR\_FD. The hollow rectangle represents a feature point descriptor. Note that  colors correspond to facial regions.}
\label{fig:FFR_FD}
\end{figure*}

\begin{figure*}[h]
  \centering
  \includegraphics[width=\linewidth]{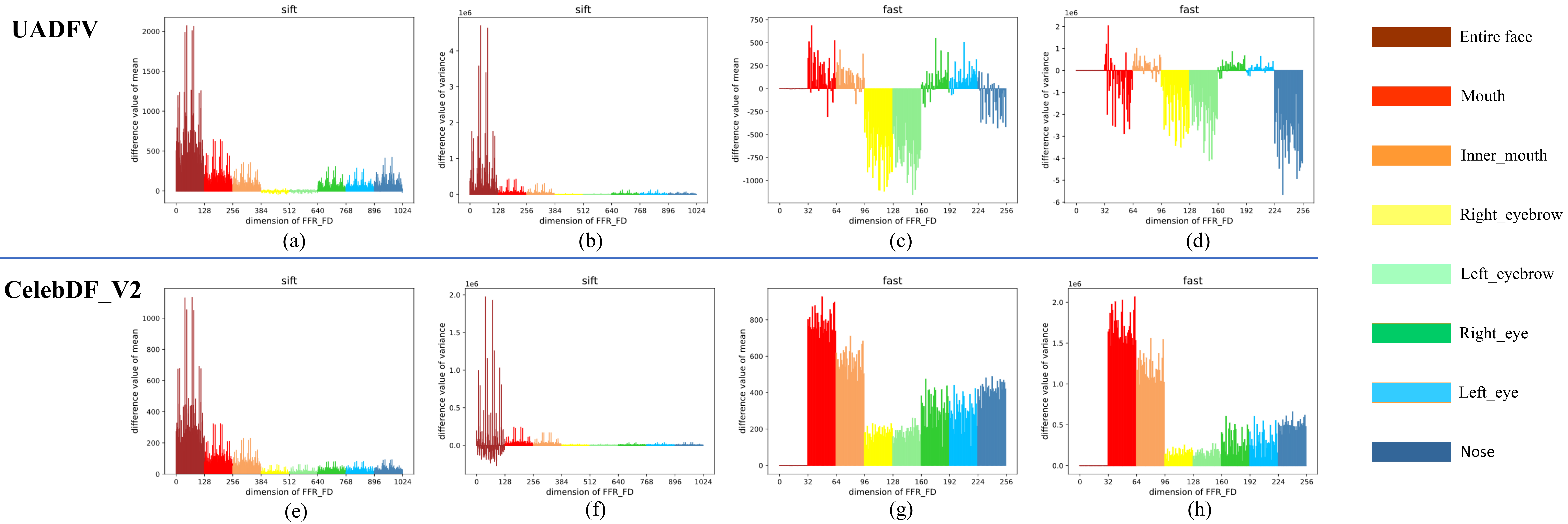}
  \caption{Differences of FFR\_FD between real and fake faces. Results are differences between statistical indicators of real and fake images  in each dimension of FFR\_FD, including the \textit{mean} and \textit{variance}, and the difference value is used as the ordinate. The feature point algorithm used to construct FFR\_FD is marked above each graph. Images are from the UADFV (above) and CelebDF\_V2 (below) training sets.}
\label{fig:FFR_FD_diff}
\end{figure*}

We divide the face into eight regions--entire face, mouth, inner\_mouth, right\_eyebrow, left\_eyebrow, right\_eye, left\_eye, and nose--where the lack of feature points is further magnified, and the descriptors of feature points from the same region have certain similarities. We detect feature points in these eight regions for a face image, and obtain corresponding descriptors. Assuming that $N_{r}$ feature points are detected in  region  $r$, and the descriptor of each feature point  $des$ is $d$-dimensional, we add the descriptors of the $N_{r}$ feature points according to the dimensions to obtain the feature descriptor of the local facial region $r$, i.e.,
\begin{equation}
FD_{r}[i]=\Sigma_{N_{r}}\mathrm{des}[i], \quad i=0,1, \ldots, d-1.
\end{equation} $FD_{r}$ is a vector that  reduces the dimensionality of the feature from $N_{r}$ x $d$ to $1$ x $d$. Note that if we do not average $FD_{r}$ (i.e., we do not divide by $N_{r}$), information about the number of feature points is introduced. $FD_{r}$ accumulates the gradient information (for SIFT and A-KAZE), intensity difference (for FAST\&BRIEF and ORB), or wavelet response (for SURF) of all of the feature point descriptors in the specified facial region. If no feature points are detected in a region, then $FD_{r}$ of that region is populated with a $d$-dimensional vector of $0$.

We concatenate   $FD_{r}$ of the eight regions in sequence to obtain an ($8$ x $d$)-dimensional vector called a fused facial region-feature descriptor (FFR\_FD), as shown in Figure \ref{fig:FFR_FD} and Algorithm \ref{FFR_FD_AL} . FFR\_FD has the following characteristics: 1) It integrates the feature descriptions of all subdivided facial regions; 2) Its calculations are  efficient. FFR\_FD can be constructed by combining any feature point detector and descriptor, so its distinguishability and speed  benefit from the feature point algorithm; 3) It can control whether it is affected by the number of feature points; and 4) Only a low-dimensional vector is used to describe the facial feature information, which significantly reduces the dimension compared to the original face image (256 x 256) or feature descriptors ($N$ x $d$), as shown in Table \ref{tab:des}.

\begin{table}[h]
  \caption{Dimensions of   feature point descriptors and FFR\_FD.}
  \label{tab:des}
  \centering
  \resizebox{\linewidth}{!}{%
  \begin{tabular}{c|c|c} 
\hline
\textbf{Algorithm} & \textbf{Descriptor Dimensions} & \textbf{FFR\_FD Dimensions}  \\ 
\hline\hline
SIFT               & 128                            & 1024 (8 x 128)                         \\ 
\hline
SURF               & 64                             & 512 (8 x 64)                          \\ 
\hline
FAST\&BRIEF          & 32                             & 256 (8 x 32)                          \\ 
\hline
ORB                & 32                             & 256 (8 x 32)                          \\ 
\hline
A-KAZE             & 61                             & 488 (8 x 61)                          \\
\hline
\end{tabular}
}
\end{table}

\subsection{Differences in FFR\_FD between real and fake faces}
If the number of  real images in a training set is $N_{R}$, then $N_{R}$ FFR\_FDs are generated. We calculate the means and variances of $N_{R}$ FFR\_FDs on the 8 x $d$ dimensions, and similarly for fake images. In a given dimension, we subtract the statistical results of the fake images from those of real ones to obtain  the statistical differences between the real and  fake FFR\_FDs in each dimension, as shown in Figure \ref{fig:FFR_FD_diff} (without averaging $FD_{r}$). In the UADFV, the mean values of FFR\_FDs constructed by the SIFT detector-descriptor for real faces are larger than those of fake faces in all dimensions, especially in the entire face and mouth regions, as shown in Figure \ref{fig:FFR_FD_diff}(a). Note that SIFT descriptors are based on gradients; hence the gradients around the real faces' feature points are stronger than those of the fake faces. Figure \ref{fig:FFR_FD_diff}(b) shows that the real FFR\_FDs have a greater variance in the entire face, i.e., the gradients' fluctuation is more significant. Feature points detected by FAST are described by BRIEF. In Figure \ref{fig:FFR_FD_diff}(c), in the mouth, Inner\_mouth, right\_eye, and left\_eye regions, the mean values of the real faces are larger, i.e., the intensity differences used to construct the BRIEF descriptor are greater, and the  opposite is true in the right\_eyebrow, left\_eyebrow, and nose regions. Figure \ref{fig:FFR_FD_diff}(d)  illustrates that the real faces have greater intensity difference fluctuations, especially in the mouth and Inner\_mouth. We similarly  analyze the other datasets, such as CelebDF\_V2, as shown at the bottom of Figure \ref{fig:FFR_FD_diff}. We conclude that real face images and DeepFakes are fairly different in FFR\_FD, and the specific distribution is affected by the feature point detector-descriptor.

\subsection{Use of only FFR\_FD to detect DeepFakes}

\begin{figure}[htb]
  \centering
  \includegraphics[width=\linewidth]{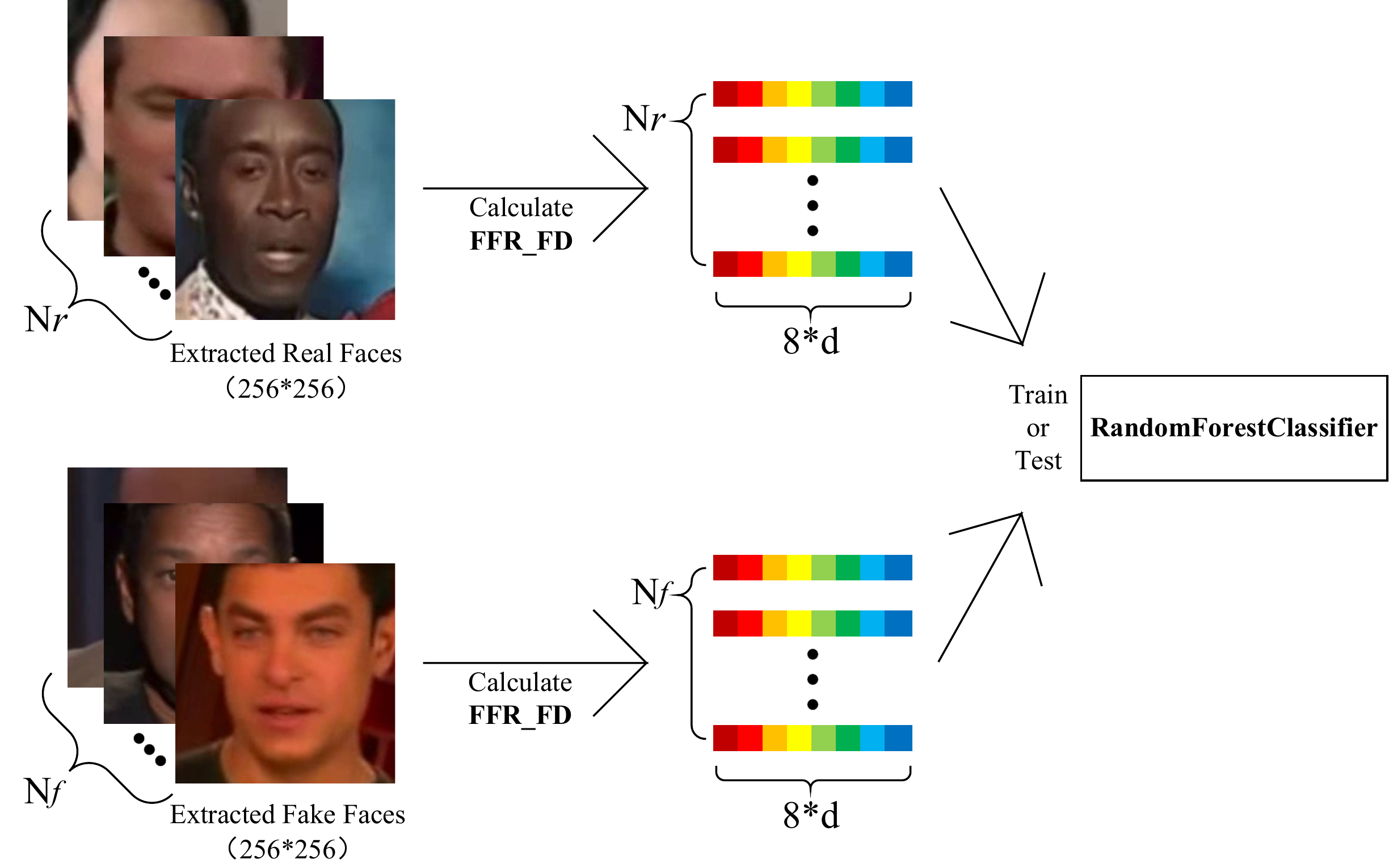}
  \caption{Simple detection pipeline.}
\label{fig:pipeline}
\end{figure}

As shown in Figure \ref{fig:pipeline}, we use FFR\_FD as the features extracted from original images to train the random forest classifier. 
We also tried LR, SVM, ELM and MLP, and found that the performance of random forest is the best in our experiments, given that it does not easily overfit. Random forest works well with   high-dimensional features, has a strong adaptability to features and a fast training speed, which is compatible with the characteristics of FFR\_FD. Besides, random forest can measure the importance of features, which has some reference significance.

\section{Experiments and results}
We describe our experimental datasets and process. Extensive experiments demonstrate the effectiveness of our method.

\subsection{Experimental setup}
\textbf{Datasets} The first-generation DeepFake datasets are DeepfakeTIMIT (HQ and LQ) \cite{korshunov2018deepfakes}, UADFV \cite{li2018ictu}, and FF++\_DeepFake (RAW and LQ) \cite{rossler2019faceforensics++}, and the second-generation are DFD \cite{dufour2019contributing}, DFDC \cite{dolhansky2019deepfake}, and CelebDF\_V2 \cite{li2020celeb}. All were used in our experiments. Each dataset was split into 80\% of the videos as the training set and the rest as the test set. For CelebDF\_V2, we followed the officially specified label to divide the test set. All videos were then extracted into frames, with the extraction rate set proportionally  to balance the number of real and fake frames. Then we extracted faces from  frames\footnote{We used \textit{S3Fd} Detector and \textit{Fan} Aligner  \cite{Faceswap}.} and   obtained facial datasets for our experiment, as shown in Table \ref{tab:datasets}. DeepFakeTIMIT and UADFV provide fewer frames, so all frames were used. In all other data sets, we only used a small part of the dataset. The   scale was reduced to take advantage of the efficiency of FFR\_FD and evaluate whether our method required as much data as deep CNN.
\begin{table}[htb]
  \caption{Details of  datasets for  experiments.}
  \label{tab:datasets}
  \centering
  \resizebox{\linewidth}{!}{%
  \begin{tabular}{c|c|c|c|c|c|c} 
\hline
\multirow{2}{*}{\textit{Dataset~}↓\textit{ Num}→} & \multicolumn{2}{c|}{Released frames} & \multicolumn{2}{c|}{Train set} & \multicolumn{2}{c}{Test Set}  \\ 
\cline{2-7}
                                                  & Real   & Fake                        & Real  & Fake                   & Real  & Fake                  \\ 
\hline\hline
DeepfakeTIMIT(HQ)                                 & 34.0k  & 34.0k                       & 27139 & 27520                  & 6870  & 6503                  \\ 
\hline
DeepfakeTIMIT(LQ)                                 & 34.0k  & 34.0k                       & 27139 & 27520                  & 6870  & 6503                  \\ 
\hline
UADFV                                             & 17.3k  & 17.3k                       & 8139  & 8138                   & 9145  & 8812                  \\ 
\hline
FF++\_DeepFakes(raw)                              & 509.9k & 509.9k                      & 29094 & 29093                  & 9556  & 9554                  \\ 
\hline
FF++\_DeepFakes(c40)                              & 509.9k & 509.9k                      & 13381 & 13376                  & 3120  & 3118                  \\ 
\hline
DFD                                               & 315.4k & 2,242.7k                    & 31324 & 30018                  & 10441 & 10800                 \\ 
\hline
DFDC                                              & 488.4k & 1,783.3k                    & 18588 & 19413                  & 4500  & 4797                  \\ 
\hline
CelebDF\_V2                                       & 225.4k & 2,116.8k                    & 38330 & 37144                  & 9968  & 9681                  \\
\hline
\end{tabular}%
}
\end{table}

\noindent\textbf{Implementation Details} We used the feature detector-descriptors of SIFT, SURF, FAST\&BRIEF, ORB, and A-KAZE to construct FFR\_FD for comparison. To investigate the influence of quantity information of feature points for detection, we   compared the detection results of FFR\_FD obtained both by averaging and not averaging $FD_{r}$. The GINI index was used as the criterion of random forest. We tried 200, 500, and 800 decision trees, and 500 was the best compromise of detection performance and efficiency. Other parameters followed Scikit-learn's default \cite{pedregosa2011scikit}. Our code is available at github \href{https://github.com/wolo-wolo/FFR_FD-Effective-and-Fast-Detection-of-DeepFakes-Based-on-Feature-Point-Defects.git}{here}.

An Nvidia GeForce RTX 2080 Ti GPU was used to extract the faces, and other experiments were performed on a consumer-level Intel Core i7-9700K CPU with 32 GB RAM.

\subsection{Results and analysis}

\begin{table*}
  \caption{AUC (\%) scores of our method and other detection methods on DeepFake datasets. The best results among all methods and between FFR\_FD are highlighted with \textbf{bold} and red, respectively.}
  \label{tab:results}
  \centering
  \scalebox{0.8}{
  \begin{tabular}{c|c|c|c|c|c|c|c|c|c|c} 
\hline
{\cellcolor[rgb]{0.753,0.753,0.753}}\textit{AUC (\%)}  & \multicolumn{2}{c|}{{\cellcolor[rgb]{0.753,0.753,0.753}}\textit{Datasets }→}      & \multirow{2}{*}{DT (HQ)}           & \multirow{2}{*}{DT (LQ)}                     & \multirow{2}{*}{UADFV}                      & \multicolumn{2}{c|}{FF++\_DF}                                           & \multirow{2}{*}{DFD}                         & \multirow{2}{*}{DFDC}                        & \multirow{2}{*}{CelebDF\_V2}                  \\ 
\cline{1-1}\cline{7-8}
{\cellcolor[rgb]{0.753,0.753,0.753}}\textit{Methods }↓ & \multicolumn{2}{c|}{{\cellcolor[rgb]{0.753,0.753,0.753}}\textit{FFR\_FD types} ↓} &                                    &                                              &                                             & (raw)                              & (c40)                              &                                              &                                              &                                               \\ 
\hline\hline
\multirow{10}{*}{Ours}                                 & \multirow{2}{*}{SIFT}      & ave                                                  & 60.8                               & 97.3                                         & 96.9                                        & 84.4                               & 79.2                               & 81.2                                         & 81.5                                         & 81.6                                          \\
                                                       &                            & no\_ave                                              & 60.5                               & 97.1                                         & 92.4                                        & 84.4                               & 79.3                               & 84.3                                         & 80.5                                         & 81.7                                          \\ 
\cline{2-11}
                                                       & \multirow{2}{*}{SURF}      & ave                                                  & 44.9                               & 86.1                                         & 97.1                                        & 78.9                               & 74.9                               & 60.4                                         & {\cellcolor[rgb]{1,0.49,0.49}}\textbf{88.3 } & {\cellcolor[rgb]{1,0.49,0.49}}\textbf{82.2 }  \\
                                                       &                            & no\_ave                                              & 63.0                               & 84.8                                         & 92.8                                        & 79.5                               & 75.2                               & 66.0                                         & 84.6                                         & 81.5                                          \\ 
\cline{2-11}
                                                       & \multirow{2}{*}{FAST\&BRIEF} & ave                                                  & 84.0                               & 99.6                                         & 96.5                                        & 89.6                               & 81.1                               & 77.3                                         & 80.0                                         & 80.5                                          \\
                                                       &                            & no\_ave                                              & {\cellcolor[rgb]{1,0.49,0.49}}85.1 & {\cellcolor[rgb]{1,0.49,0.49}}\textbf{99.9 } & 93.2                                        & {\cellcolor[rgb]{1,0.49,0.49}}92.3 & {\cellcolor[rgb]{1,0.49,0.49}}81.6 & {\cellcolor[rgb]{1,0.49,0.49}}\textbf{85.2 } & 69.9                                         & 78.0                                          \\ 
\cline{2-11}
                                                       & \multirow{2}{*}{ORB}       & ave                                                  & 50.7                               & 98.6                                         & 90.5                                        & 81.5                               & 77.5                               & 69.8                                         & 78.6                                         & 76.1                                          \\
                                                       &                            & no\_ave                                              & 65.4                               & 98.1                                         & 87.4                                        & 84.9                               & 77.6                               & 76.3                                         & 67.8                                         & 74.1                                          \\ 
\cline{2-11}
                                                       & \multirow{2}{*}{A-KAZE}    & ave                                                  & 41.8                               & 78.1                                         & {\cellcolor[rgb]{1,0.49,0.49}}\textbf{97.8} & 81.1                               & 77.5                               & 56.3                                         & 82.0                                         & 79.9                                          \\
                                                       &                            & no\_ave                                              & 48.2                               & 86.7                                         & 96.3                                        & 79.2                               & 74.0                               & 62.2                                         & 76.6                                         & 79.5                                          \\ 
\hline
\multicolumn{3}{c|}{Two-stream NN \cite{zhou2017two}}                                                                                                         & 73.5                               & 83.5                                         & 85.1                                        & 70.1                               & -                                  & 52.8                                         & 61.4                                         & 53.8                                          \\ 
\hline
\multicolumn{3}{c|}{Meso-4 \cite{afchar2018mesonet}}                                                                                                                & 68.4                               & 87.8                                         & 84.3                                        & 84.7                               & -                                  & 76.0                                         & 75.3                                         & 54.8                                          \\
\multicolumn{3}{c|}{MesoInception-4}                                                                                                       & 62.7                               & 80.4                                         & 82.1                                        & 83.0                               & -                                  & 75.9                                         & 73.2                                         & 53.6                                          \\ 
\hline
\multicolumn{3}{c|}{HeadPose \cite{yang2019exposing}}                                                                                                              & 53.2                               & 55.1                                         & 89.0                                        & 47.3                               & -                                  & 56.1                                         & 55.9                                         & 54.6                                          \\ 
\hline
\multicolumn{3}{c|}{FWA \cite{li2018exposing}}                                                                                                                   & 93.2                               & 99.9                                         & 97.4                                        & 80.1                               & -                                  & 74.3                                         & 72.7                                         & 53.9                                          \\ 
\hline
\multicolumn{3}{c|}{VA\_mLP \cite{matern2019exploiting}}                                                                                                               & 62.1                               & 61.4                                         & 70.2                                        & 66.4                               & -                                  & 69.1                                         & 61.9                                         & 55.0                                          \\
\multicolumn{3}{c|}{VA-LogReg}                                                                                                             & 77.3                               & 77.0                                         & 54.0                                        & 78.0                               & -                                  & 77.2                                         & 66.2                                         & 55.1                                          \\ 
\hline
\multicolumn{3}{c|}{Xception-raw \cite{rossler2019faceforensics++}}                                                                                                          & 54.0                               & 56.7                                         & 80.4                                        & \textbf{99.7 }                     & -                                  & 53.9                                         & 49.9                                         & 48.2                                          \\
\multicolumn{3}{c|}{Xception-c40}                                                                                                          & 70.5                               & 75.8                                         & 83.6                                        & 95.6                               & -                                  & 65.8                                         & 69.7                                         & 65.5                                          \\ 
\hline
\multicolumn{3}{c|}{Multi-task \cite{nguyen2019multi}}                                                                                                            & 65.8                               & 62.2                                         & 55.3                                        & 76.3                               & -                                  & 54.1                                         & 53.6                                         & 54.3                                          \\ 
\hline
\multicolumn{3}{c|}{Capsule \cite{nguyen2019use}}                                                                                                               & 74.4                               & 78.4                                         & 61.3                                        & 96.6                               & -                                  & 64.0                                         & 53.3                                         & 57.5                                          \\ 
\hline
\multicolumn{3}{c|}{DSP-FWA \cite{he2015spatial}}                                                                                                               & \textbf{99.7 }                     & \textbf{99.9 }                               & 97.7                                        & 93.0                               & -                                  & 81.1                                         & 75.5                                         & 64.6                                          \\
\hline
\end{tabular}
}
\end{table*}

\begin{table*}[htb]
  \caption{AUC (\%) scores on other datasets for classifier only trained with DeepFakeTIMIT (HQ)}
  \label{tab:gen}
  \centering
  \resizebox{\linewidth}{!}{%
  \begin{tabular}{c|c|c|c|c|c|c|c|c|c} 
\hline
\multicolumn{2}{c|}{\multirow{2}{*}{\textit{Methods}}} & \multirow{2}{*}{\textit{Train}} & \multicolumn{7}{c}{\textit{Test}}                                           \\ 
\cline{4-10}
\multicolumn{2}{c|}{}                                  &                                 & DT (LQ) & UADFV & FF++\_DF(RAW) & FF++\_DF(LQ) & DFD  & DFDC & CelebDF\_V2  \\ 
\hline \hline
\multirow{2}{*}{FAST\&BRIEF} & ave                       & \multirow{5}{*}{DT (HQ)}        & 25.8    & \textbf{67.7}  & 50.0          & 48.6         & 52.3 & 44.6 & 39.9         \\ 
\cline{2-2}\cline{4-10}
                           & no\_ave                   &                                 & 98.5    & \textbf{67.7}  & \textbf{80.0}         & \textbf{73.1}         & \textbf{67.8} & 52.8 & \textbf{68.8}         \\ 
\cline{1-2}\cline{4-10}
\multicolumn{2}{c|}{ResNet50}                          &                                 & \textbf{100.0}   & 48.5  & 49.4          & 53.2         & 53.2 & 53.4 & 51.3         \\ 
\cline{1-2}\cline{4-10}
\multicolumn{2}{c|}{Xception}                          &                                 & \textbf{100.0}   & 50.3  & 50.0          & 50.0         & 56.4 & 51.6 & 50.2         \\ 
\cline{1-2}\cline{4-10}
\multicolumn{2}{c|}{EfficientNetB0}                    &                                 & \textbf{100.0}   & 50.0  & 50.0          & 50.0         & 54.1 & \textbf{54.1} & 53.2         \\
\hline
\end{tabular}%
}
\end{table*}

Using the frame-level ROC-AUC score as the metric, Table \ref{tab:results} reports the test results of random forest with 500 subtrees and provides some baselines for comparison. Among five detector-descriptors, the overall detection performance of FFR\_FD extracted by the FAST detector combined with the BRIEF descriptor was more effective and stable. As described in Section \ref{sec:FP_num}, the number of feature points of fake faces detected by FAST was reduced significantly. However,   ORB with the same significant reduction in the number of feature points did not have ideal performance. ORB is based on FAST\&BRIEF to improve   scale- and rotation-invariance, but all faces were cropped and aligned in the DeepFake detection task. Scale- and rotation-invariance are not required, but will weaken the discriminability of FFR\_FD. Except for FAST\&BRIEF,  descriptor-detectors were designed with these two properties in mind. As shown in Table \ref{tab:results}, the introduction of feature point quantity information did not improve AUC in all of the cases, indicating that the feature point descriptors themselves are sufficiently distinguishable. Comparing the detection results of the RAW and c40 (LQ) versions of FF++\_DF, our method was affected slightly   by strong compression, but was still acceptable. On the DT (LQ), UADFV, and the extremely challenging DFDC and CelebDF\_V2 datasets, our detection performance was ahead of all   methods and was  competitive on other datasets.

It is worth mentioning the following: 1) FFR\_FD is just a vector of length 8 x $d$ (shown in Table \ref{tab:des}), and the length   is only $256$ for one face if constructed based on FAST\&BRIEF; 2) Driven by large-scale training data or features, models such as Two-stream (a two-stream CNN), FWA/DSP-FWA (ResNet-50), Xception, Multi-task (CNN-based), Capsule (VGG19-based), or current methods  require expensive GPU resources to train large and complex CNNs. We only used a consumer-level CPU to train the random forest, with comparable or even better results; 3) Methods that also use simple machine learning models to capture specific features, such as HeadPose (SVM) and VA (multilayer perceptron or logistic regression), are far from satisfactory because captured superficial visual artifacts  are not refined enough, resulting in limited detection capabilities and easy repair. However, deep generative models are challenged to generate DeepFakes with enough feature points, i.e., DeepFakes lack fine-grained information (e.g., gradient, intense differences)  for feature point detection. This makes our approach SOTA; and 4) FFR\_FD benefits from   feature point detector-descriptors such as speed and reliability, and FAST\&BRIEF are the first choices. In summary, our method has advantages in speed, performance, storage, and computational costs.

\subsection{Generalization test and results}

We only used the random forest with 200 subtrees trained by DeepfakeTIMIT (HQ) as the classifier, and all of the other datasets were used as test sets for generalization evaluation. It is an arduous task because DeepfakeTIMIT (HQ) is a first-generation DeepFake dataset, with few frames available. Subsequent datasets comprehensively improved both the quantity and quality of DeepFake videos. We believe we are the first to test generalization in this challenging setting. ResNet50 \cite{he2016deep}, Xception \cite{chollet2017xception}, and EfficientNet-B0 \cite{tan2019efficientnet} are state-of-the-art DeepFake detection methods, so we chose them as baselines. We directly used baseline models provided by Keras \cite{Keras}, and only modified the fully-connected layers for binary classification. The results of baselines and FFR\_FD (constructed by FAST\&BRIEF) are shown in Table \ref{tab:gen}. Comparing \textit{ave} and \textit{no\_ave}, the introduction of feature point quantity information can  improve  generalizability, benefiting from the lack of feature points in most DeepFake datasets. Results show that our method has considerable generalization ability, while baseline methods are limited to the training set and suffer from overfitting. FFR\_FD  constructed only from DeepfakeTIMIT (HQ) was also effective to detect other datasets, except DFDC. DFDC involves multiple synthesis algorithms, and FFR\_FD constructed from DFDC has little similarity with DeepFakeTIMIT.

\subsection{Comparison with other state-of-the-arts on CelebDF\_V2}

\begin{figure}[h!]
  \centering
  \includegraphics[width=\linewidth]{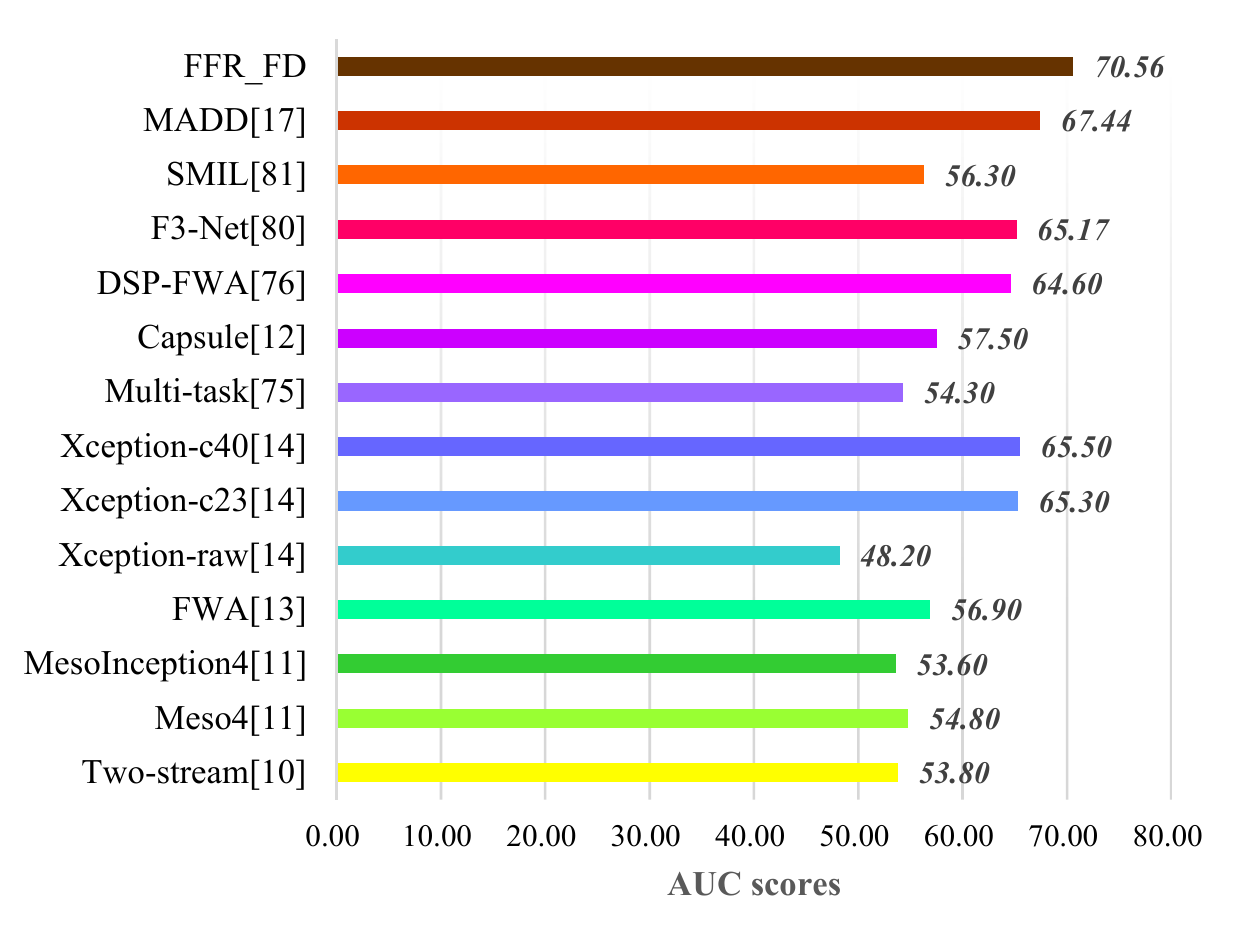}
  \caption{The AUC(\%) score of Cross-dataset evaluation on CelebDF\_V2.}
\label{fig:CB}
\end{figure}

\begin{figure*}[h!]
  \centering
  \includegraphics[width=\textwidth]{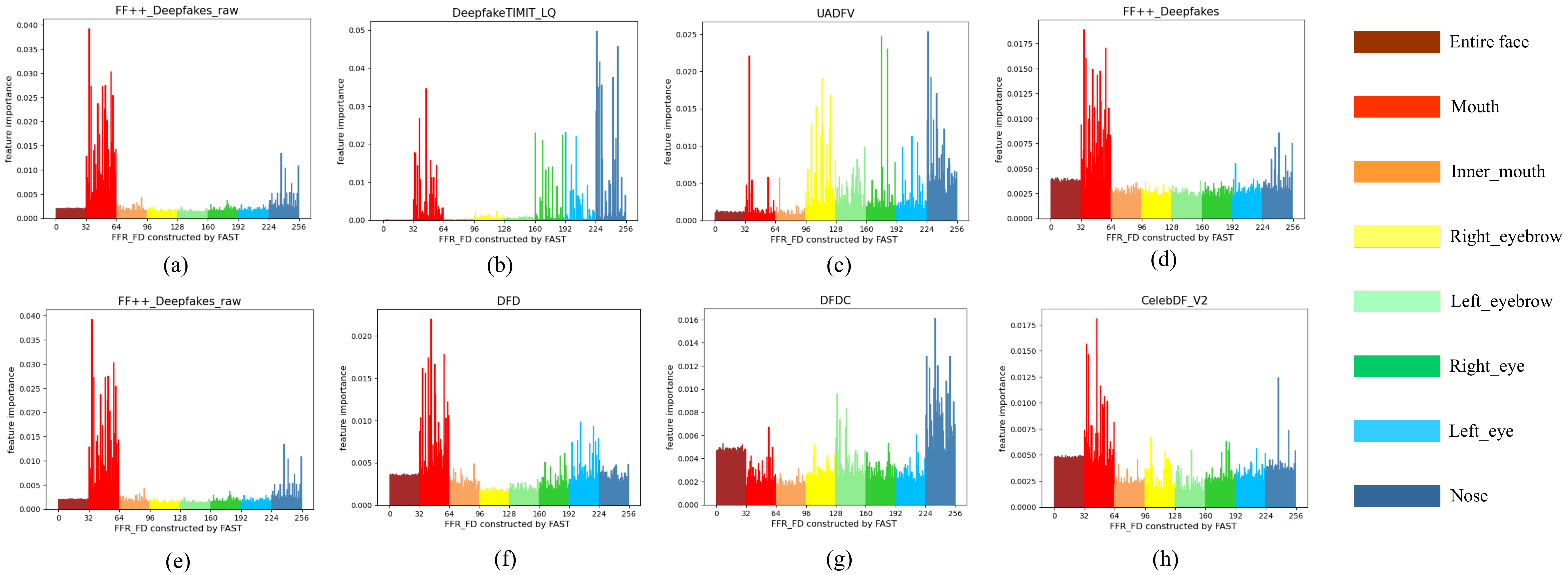}
  \caption{Feature importance of FFR\_FD for random forest classification. Results come from FFR\_FD constructed by FAST detector and BRIEF descriptor. The dataset is marked above each subgraph.}
\label{fig:fi}
\end{figure*}

CelebDF\_V2 dataset improves the forgery quality and significantly reduces visible artifacts compared to the FF++ benchmark. Following convention \cite{li2020celeb, zhao2021multi}, we train the model on FF++ while test it on CelebDF\_V2 (i.e., \textit{no\_ave} FFR\_FD construct from FF++ are used to train the model), for evaluating the transferability of our method. The frame-level average AUC score is shown in Figure \ref{fig:CB}. Even compared with the recently proposed MADD\cite{zhao2021multi}, F3-Net\cite{qian2020thinking}, and SMIL\cite{li2020sharp}, FFR\_FD further improves the metrics by 3.12\%, 6.27\%, and 14.26\%, respectively. The results demonstrate our method achieves state-of-the-art generalizability. As shown in Table \ref{tab:FP_NUM_datasets}, in the extremely challenging CelebDF\_V2, the facial regions of forged faces equally lack sufficient feature points, though the synthesis algorithms are improved. Moreover, the \textit{feature importances} of FFR\_FD focused by random forest actually show a nearly same distribution on FF++ and CelebDF\_V2, as shown in Figure \ref{fig:fi} (d) and (h). These provide new insights into the success of performance gains.

\subsection{Feature importances}

\begin{figure}[h!]
  \centering
  \includegraphics[width=\linewidth]{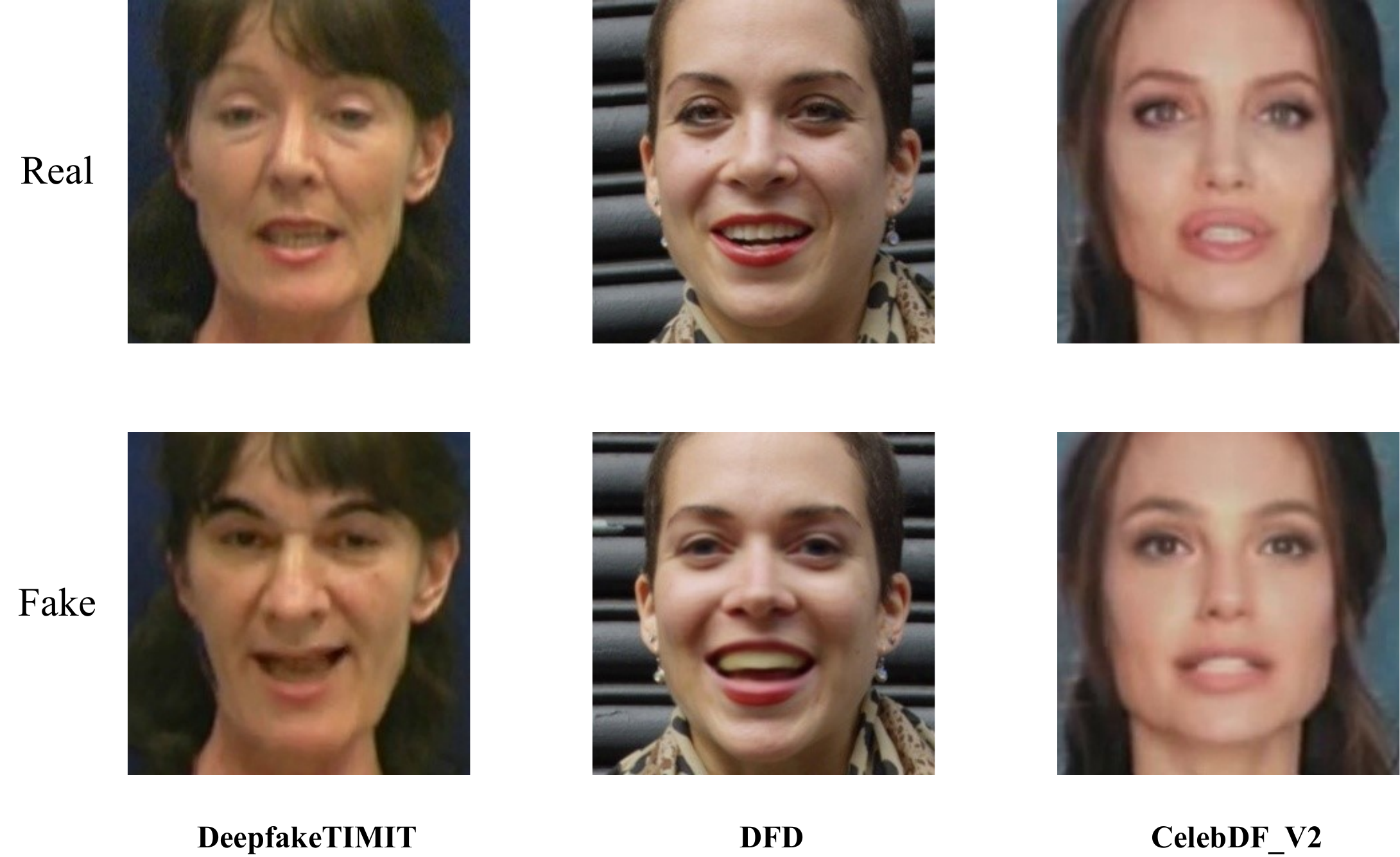}
  \caption{Real and corresponding DeepFake faces.}
\label{fig:de}
\end{figure}

We divide the face into eight regions, entire face, mouth, inner\_mouth, right\_eyebrow, left\_eyebrow, right\_eye, left\_eye and nose to construct FFR\_FD. Are these facial regions both informative for DeepFake detection? To answer this question, we output the feature importance (also known as the Gini importance) of random forest. Feature importance is computed as the normalized total reduction of the criterion brought by that feature. Figure \ref{fig:fi} shows the feature importance of FFR\_FD constructed by FAST\&BRIEF on different datasets. On DeepfakeTIMIT, FF++\_DeepFakes, DFD, and CelebDF\_V2, the mouth region is the crucial feature for identification. As shown in Figure \ref{fig:de}, the forged faces in these datasets have artifact defects on the mouth, which causes the feature point descriptors of mouth to struggle with real ones. Except for the mouth, Figure \ref{fig:fi} shows that most facial regions contribute to the random forest classification with FFR\_FD. Considering that when detecting DeepFake in the wild, we should not use prior knowledge of defects in some regions of a specific dataset.

\section{The efficiency of FFR\_FD}

\begin{figure}[h]
  \centering
  \includegraphics[width=.9\linewidth]{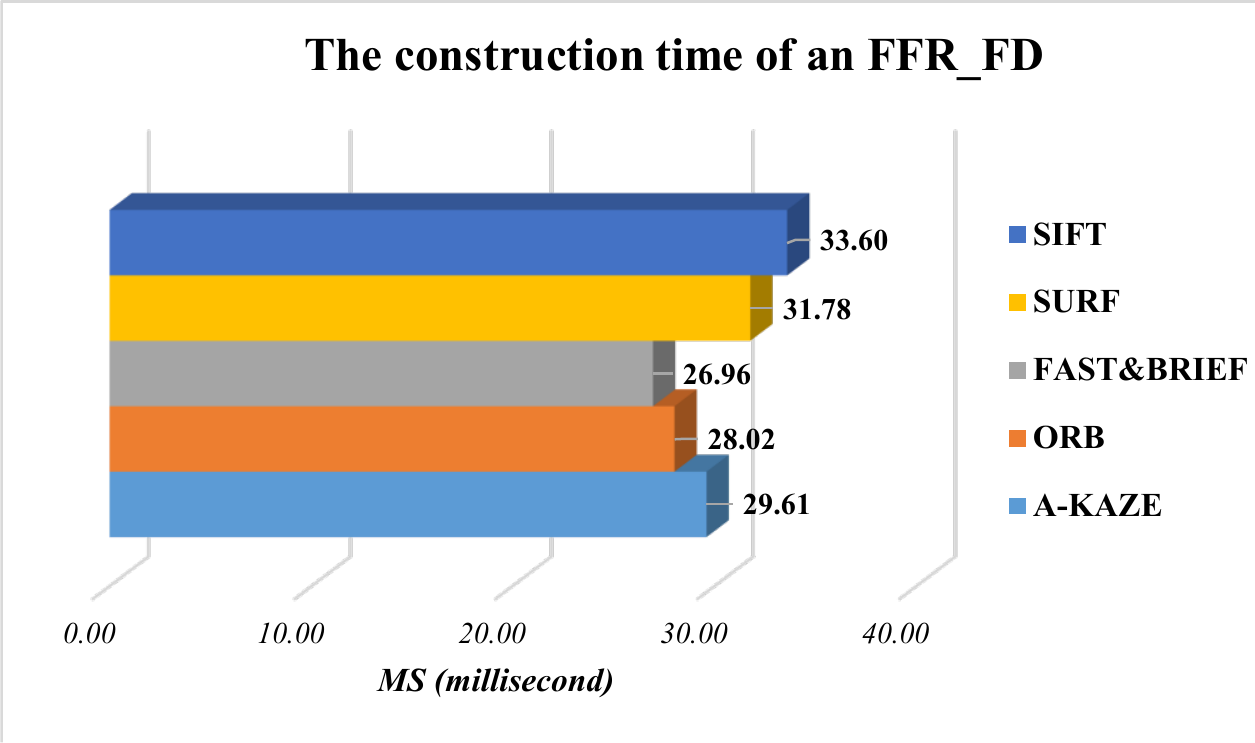}
  \caption{The average construction time of an FFR\_FD.
}
\label{ft}
\end{figure}

We conduct FFR\_FD construction and random forest training on a consumer-level Intel Core i7-9700K CPU with 32 GB RAM. Figure \ref{ft} shows the average calculation time of constructing an FFR\_FD from a face image. The quantitative comparison shows that the computational efficiency of feature detection-descriptor for computing FFR\_FD is: \textit{FAST}\&\textit{BRIEF $>$ ORB $>$ A-KAZE $>$ SURF $>$ SIFT}. As we expected, it depends on the calculation speed of the feature point algorithm.

\begin{figure}[h]
  \centering
  \includegraphics[width=.9\linewidth]{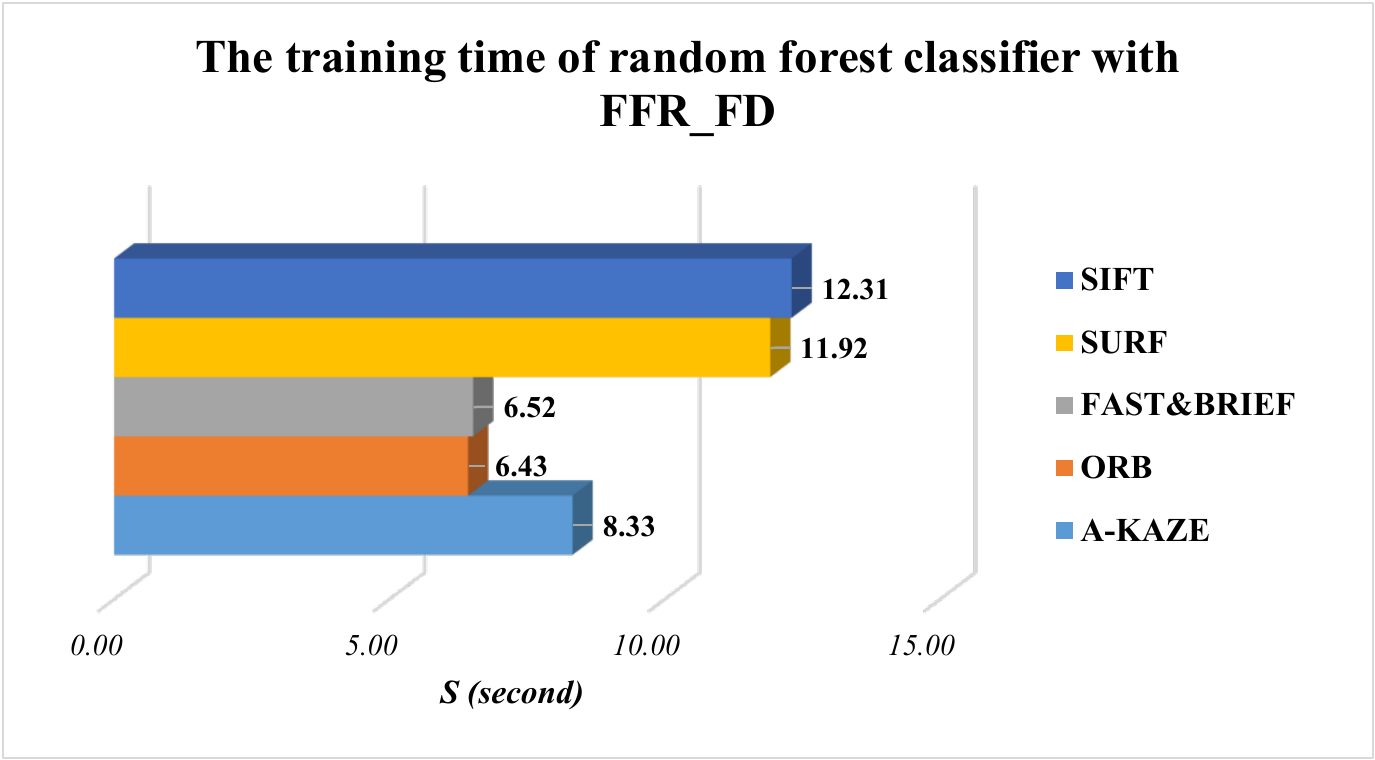}
  \caption{The average training time of random forest classifier with FFR\_FD.}
\label{fig:tt}
\end{figure}

We randomly use 10K FFR\_FD (i.e., extracted from 10K images) to train a random forest classifier with 500 subtrees and compare the average training time in Figure \ref{fig:tt}. It can be seen that the training time is positively related to the dimension of FFR\_FD. In our experimental environment, if 100K face images are used and FFR\_FD is constructed from FAST\&BRIEF, it only takes about 45 minutes to complete the construction and training process. Even with the help of an advanced GPU, driving 100K face images to train deep neural networks is more time-consuming than our approach.

\section{Limitation and conclusion}
We presented FFR\_FD, a vector representation for DeepFake detection, which can be constructed from different facial regions in combination with various feature descriptors. Inspired by local feature detection-description algorithms to extract fine-grained features, we explored the feature points in DeepFakes. Through using five feature point detector-descriptors, SIFT, SURF, FAST\&BRIEF, ORB, and A-KAZE,  the experimental results indicate current DeepFake faces lack a sufficient number of feature points. Without the need for powerful GPUs, we trained the random forest classifier with FFR\_FD. Experimental results showed that our approach can achieve state-of-the-art detection performance while considering efficiency and generalization.

FFR\_FD relies heavily on feature point detector-descriptors, but current algorithms are not specifically designed for DeepFake detection tasks, given that they must compromise between distinguishability and invariance. In future work, we would like to design a discriminative feature descriptor for face forensics.

\section*{CRediT authorship contribution statement}
\textbf{Gaojian Wang:} Conceptualization, Methodology, Software, Investigation, Writing - Original Draft. \textbf{Qian Jiang:} Funding acquisition, Formal analysis, Writing - Review \& Editing, Data Curation. \textbf{Xin Jin:} Software, Validation, Visualization, Resources.  \textbf{Xiaohui Cui:}  Writing - Review \& Editing, Project administration, Supervision, Resources.

\section*{Declaration of competing interes}
The authors declare that they have no known competing financial
interests or personal relationships that could have appeared
to influence the work reported in this paper.

\section*{ACKNOWLEDGMENT}
This research was funded by the National Natural Science Foundation of China (No. 62002313), Key Areas Research Program of Yunnan Province in China (202001BB050076), Key Laboratory in Software Engineering of Yunnan Province in China (2020SE408), Postdoctoral Science Foundation of Yunnan Province in China.

\appendix

\section{Normalization of FFR\_FD}
In our experiments, FFR\_FD does not need to be normalized, given that feature point descriptors have been normalized. SIFT, BRIEF, ORB, and A-KAZE are all normalized at [0, 255], and SURF descriptor based on the sum of the Haar wavelet response is at [-1, 1]. We construct $FD_{r}$ by accumulating the same type descriptors in the facial region $r$, and the resulting FFR\_FDs have dimensional unification. On the other hand, compare with LR (logistic regression), SVM (support vector machine), and MLP (multilayer perceptron), the random forest is the best classifier according to our test results. It does not need to standardize the features given the decision tree does not involve distance measurement.


 \bibliographystyle{elsarticle-num} 
 \bibliography{cas-refs}

\begin{thebibliography}{10}
\expandafter\ifx\csname url\endcsname\relax
  \def\url#1{\texttt{#1}}\fi
\expandafter\ifx\csname urlprefix\endcsname\relax\def\urlprefix{URL }\fi
\expandafter\ifx\csname href\endcsname\relax
  \def\href#1#2{#2} \def\path#1{#1}\fi

\bibitem{goodfellow2014generative}
I.~J. Goodfellow, J.~Pouget-Abadie, M.~Mirza, B.~Xu, D.~Warde-Farley, S.~Ozair,
  A.~Courville, Y.~Bengio, Generative adversarial networks, arXiv preprint
  arXiv:1406.2661 (2014).

\bibitem{kingma2013auto}
D.~P. Kingma, M.~Welling, Auto-encoding variational bayes, arXiv preprint
  arXiv:1312.6114 (2013).

\bibitem{Avatarify}
Avatarify,
  \url{https://apps.apple.com/us/app/avatarify-ai-face-animator/id1512669147}.
  Accessed:2021-03.

\bibitem{Reface}
Reface,
  \url{https://apps.apple.com/us/app/reface-face-swap-videos/id1488782587}.
  Accessed:2021-03.

\bibitem{BBC}
B.~Bitesize, “deepfakes: What are they and why would i make one?”,
  \url{https://www.bbc.co.uk/bitesize/articles/zfkwcqt}. Accessed:2021-03.

\bibitem{VICE}
S.~Cole, “we are truly fucked: Everyone is making ai-generated fake porn
  now”,
  \url{https://www.vice.com/en/article/bjye8a/reddit-fake-porn-app-daisy-ridley}.
  Accessed:2021-03.

\bibitem{agarwal2019protecting}
S.~Agarwal, H.~Farid, Y.~Gu, M.~He, K.~Nagano, H.~Li, Protecting world leaders
  against deep fakes., in: CVPR Workshops, 2019, pp. 38--45.

\bibitem{yang2019exposing}
X.~Yang, Y.~Li, S.~Lyu, Exposing deep fakes using inconsistent head poses, in:
  ICASSP 2019-2019 IEEE International Conference on Acoustics, Speech and
  Signal Processing (ICASSP), IEEE, 2019, pp. 8261--8265.

\bibitem{li2018ictu}
Y.~Li, M.-C. Chang, S.~Lyu, In ictu oculi: Exposing ai created fake videos by
  detecting eye blinking, in: 2018 IEEE International Workshop on Information
  Forensics and Security (WIFS), IEEE, 2018, pp. 1--7.

\bibitem{zhou2017two}
P.~Zhou, X.~Han, V.~I. Morariu, L.~S. Davis, Two-stream neural networks for
  tampered face detection, in: 2017 IEEE Conference on Computer Vision and
  Pattern Recognition Workshops (CVPRW), IEEE, 2017, pp. 1831--1839.

\bibitem{afchar2018mesonet}
D.~Afchar, V.~Nozick, J.~Yamagishi, I.~Echizen, Mesonet: a compact facial video
  forgery detection network, in: 2018 IEEE International Workshop on
  Information Forensics and Security (WIFS), IEEE, 2018, pp. 1--7.

\bibitem{nguyen2019use}
H.~H. Nguyen, J.~Yamagishi, I.~Echizen, Use of a capsule network to detect fake
  images and videos, arXiv preprint arXiv:1910.12467 (2019).

\bibitem{li2018exposing}
Y.~Li, S.~Lyu, Exposing deepfake videos by detecting face warping artifacts,
  arXiv preprint arXiv:1811.00656 (2018).

\bibitem{rossler2019faceforensics++}
A.~Rossler, D.~Cozzolino, L.~Verdoliva, C.~Riess, J.~Thies, M.~Nie{\ss}ner,
  Faceforensics++: Learning to detect manipulated facial images, in:
  Proceedings of the IEEE/CVF International Conference on Computer Vision,
  2019, pp. 1--11.

\bibitem{li2020celeb}
Y.~Li, X.~Yang, P.~Sun, H.~Qi, S.~Lyu, Celeb-df: A large-scale challenging
  dataset for deepfake forensics, in: Proceedings of the IEEE/CVF Conference on
  Computer Vision and Pattern Recognition, 2020, pp. 3207--3216.

\bibitem{dang2020detection}
H.~Dang, F.~Liu, J.~Stehouwer, X.~Liu, A.~K. Jain, On the detection of digital
  face manipulation, in: Proceedings of the IEEE/CVF Conference on Computer
  Vision and Pattern Recognition, 2020, pp. 5781--5790.

\bibitem{zhao2021multi}
H.~Zhao, W.~Zhou, D.~Chen, T.~Wei, W.~Zhang, N.~Yu, Multi-attentional deepfake
  detection, arXiv preprint arXiv:2103.02406 (2021).

\bibitem{zi2020wilddeepfake}
B.~Zi, M.~Chang, J.~Chen, X.~Ma, Y.-G. Jiang, Wilddeepfake: A challenging
  real-world dataset for deepfake detection, in: Proceedings of the 28th ACM
  International Conference on Multimedia, 2020, pp. 2382--2390.

\bibitem{liu2020global}
Z.~Liu, X.~Qi, P.~H. Torr, Global texture enhancement for fake face detection
  in the wild, in: Proceedings of the IEEE/CVF Conference on Computer Vision
  and Pattern Recognition, 2020, pp. 8060--8069.

\bibitem{sun2020identifying}
X.~Sun, B.~Wu, W.~Chen, Identifying invariant texture violation for robust
  deepfake detection, arXiv preprint arXiv:2012.10580 (2020).

\bibitem{chugh2020not}
K.~Chugh, P.~Gupta, A.~Dhall, R.~Subramanian, Not made for each
  other-audio-visual dissonance-based deepfake detection and localization, in:
  Proceedings of the 28th ACM International Conference on Multimedia, 2020, pp.
  439--447.

\bibitem{zhang2019detecting}
X.~Zhang, S.~Karaman, S.-F. Chang, Detecting and simulating artifacts in gan
  fake images, in: 2019 IEEE International Workshop on Information Forensics
  and Security (WIFS), IEEE, 2019, pp. 1--6.

\bibitem{rosten2006machine}
E.~Rosten, T.~Drummond, Machine learning for high-speed corner detection, in:
  European conference on computer vision, Springer, 2006, pp. 430--443.

\bibitem{van2008visualizing}
L.~Van~der Maaten, G.~Hinton, Visualizing data using t-sne., Journal of machine
  learning research 9~(11) (2008).

\bibitem{lowe2004distinctive}
D.~G. Lowe, Distinctive image features from scale-invariant keypoints,
  International journal of computer vision 60~(2) (2004) 91--110.

\bibitem{mian2010repeatability}
A.~Mian, M.~Bennamoun, R.~Owens, On the repeatability and quality of keypoints
  for local feature-based 3d object retrieval from cluttered scenes,
  International Journal of Computer Vision 89~(2) (2010) 348--361.

\bibitem{kim2012local}
D.~Kim, S.~Rho, E.~Hwang, Local feature-based multi-object recognition scheme
  for surveillance, Engineering Applications of Artificial Intelligence 25~(7)
  (2012) 1373--1380.

\bibitem{marchand2015pose}
E.~Marchand, H.~Uchiyama, F.~Spindler, Pose estimation for augmented reality: a
  hands-on survey, IEEE transactions on visualization and computer graphics
  22~(12) (2015) 2633--2651.

\bibitem{tareen2018comparative}
S.~A.~K. Tareen, Z.~Saleem, A comparative analysis of sift, surf, kaze, akaze,
  orb, and brisk, in: 2018 International conference on computing, mathematics
  and engineering technologies (iCoMET), IEEE, 2018, pp. 1--10.

\bibitem{krig2016interest}
S.~Krig, Interest point detector and feature descriptor survey, in: Computer
  vision metrics, Springer, 2016, pp. 187--246.

\bibitem{Faceswap}
faceswap github, \url{https://github.com/deepfakes/faceswap}. Accessed:2021-03.

\bibitem{korshunov2018deepfakes}
P.~Korshunov, S.~Marcel, Deepfakes: a new threat to face recognition?
  assessment and detection, arXiv preprint arXiv:1812.08685 (2018).

\bibitem{dufour2019contributing}
N.~Dufour, A.~Gully, Contributing data to deepfake detection research, Google
  AI Blog (2019).

\bibitem{dolhansky2019deepfake}
B.~Dolhansky, R.~Howes, B.~Pflaum, N.~Baram, C.~C. Ferrer, The deepfake
  detection challenge (dfdc) preview dataset, arXiv preprint arXiv:1910.08854
  (2019).

\bibitem{calonder2010brief}
M.~Calonder, V.~Lepetit, C.~Strecha, P.~Fua, Brief: Binary robust independent
  elementary features, in: European conference on computer vision, Springer,
  2010, pp. 778--792.

\bibitem{rublee2011orb}
E.~Rublee, V.~Rabaud, K.~Konolige, G.~Bradski, Orb: An efficient alternative to
  sift or surf, in: 2011 International conference on computer vision, Ieee,
  2011, pp. 2564--2571.

\bibitem{bay2008speeded}
H.~Bay, A.~Ess, T.~Tuytelaars, L.~Van~Gool, Speeded-up robust features (surf),
  Computer vision and image understanding 110~(3) (2008) 346--359.

\bibitem{alcantarilla2011fast}
P.~F. Alcantarilla, T.~Solutions, Fast explicit diffusion for accelerated
  features in nonlinear scale spaces, IEEE Trans. Patt. Anal. Mach. Intell
  34~(7) (2011) 1281--1298.

\bibitem{karras2019style}
T.~Karras, S.~Laine, T.~Aila, A style-based generator architecture for
  generative adversarial networks, in: Proceedings of the IEEE/CVF Conference
  on Computer Vision and Pattern Recognition, 2019, pp. 4401--4410.

\bibitem{nirkin2019fsgan}
Y.~Nirkin, Y.~Keller, T.~Hassner, Fsgan: Subject agnostic face swapping and
  reenactment, in: Proceedings of the IEEE/CVF International Conference on
  Computer Vision, 2019, pp. 7184--7193.

\bibitem{karras2020analyzing}
T.~Karras, S.~Laine, M.~Aittala, J.~Hellsten, J.~Lehtinen, T.~Aila, Analyzing
  and improving the image quality of stylegan, in: Proceedings of the IEEE/CVF
  Conference on Computer Vision and Pattern Recognition, 2020, pp. 8110--8119.

\bibitem{tewari2017mofa}
A.~Tewari, M.~Zollhofer, H.~Kim, P.~Garrido, F.~Bernard, P.~Perez, C.~Theobalt,
  Mofa: Model-based deep convolutional face autoencoder for unsupervised
  monocular reconstruction, in: Proceedings of the IEEE International
  Conference on Computer Vision Workshops, 2017, pp. 1274--1283.

\bibitem{thies2016face2face}
J.~Thies, M.~Zollhofer, M.~Stamminger, C.~Theobalt, M.~Nie{\ss}ner, Face2face:
  Real-time face capture and reenactment of rgb videos, in: Proceedings of the
  IEEE conference on computer vision and pattern recognition, 2016, pp.
  2387--2395.

\bibitem{thies2019deferred}
J.~Thies, M.~Zollh{\"o}fer, M.~Nie{\ss}ner, Deferred neural rendering: Image
  synthesis using neural textures, ACM Transactions on Graphics (TOG) 38~(4)
  (2019) 1--12.

\bibitem{tolosana2020deepfakes}
R.~Tolosana, R.~Vera-Rodriguez, J.~Fierrez, A.~Morales, J.~Ortega-Garcia,
  Deepfakes and beyond: A survey of face manipulation and fake detection,
  Information Fusion 64 (2020) 131--148.

\bibitem{FakeAPP}
F.~github, \url{https://www.malavida.com/en/soft/fakeapp/}. Accessed:2021-03.

\bibitem{faceswap-GAN}
faceswap GAN~github, \url{https://github.com/shaoanlu/faceswap-GAN}.
  Accessed:2021-03.

\bibitem{ciftci2020fakecatcher}
U.~A. Ciftci, I.~Demir, L.~Yin, Fakecatcher: Detection of synthetic portrait
  videos using biological signals, IEEE Transactions on Pattern Analysis and
  Machine Intelligence (2020).

\bibitem{marra2018detection}
F.~Marra, D.~Gragnaniello, D.~Cozzolino, L.~Verdoliva, Detection of
  gan-generated fake images over social networks, in: 2018 IEEE Conference on
  Multimedia Information Processing and Retrieval (MIPR), IEEE, 2018, pp.
  384--389.

\bibitem{amerini2020exploiting}
I.~Amerini, R.~Caldelli, Exploiting prediction error inconsistencies through
  lstm-based classifiers to detect deepfake videos, in: Proceedings of the 2020
  ACM Workshop on Information Hiding and Multimedia Security, 2020, pp.
  97--102.

\bibitem{montserrat2020deepfakes}
D.~M. Montserrat, H.~Hao, S.~K. Yarlagadda, S.~Baireddy, R.~Shao,
  J.~Horv{\'a}th, E.~Bartusiak, J.~Yang, D.~Guera, F.~Zhu, et~al., Deepfakes
  detection with automatic face weighting, in: Proceedings of the IEEE/CVF
  Conference on Computer Vision and Pattern Recognition Workshops, 2020, pp.
  668--669.

\bibitem{wang2019fakespotter}
R.~Wang, F.~Juefei-Xu, L.~Ma, X.~Xie, Y.~Huang, J.~Wang, Y.~Liu, Fakespotter: A
  simple yet robust baseline for spotting ai-synthesized fake faces, arXiv
  preprint arXiv:1909.06122 (2019).

\bibitem{xie2019deephunter}
X.~Xie, L.~Ma, F.~Juefei-Xu, M.~Xue, H.~Chen, Y.~Liu, J.~Zhao, B.~Li, J.~Yin,
  S.~See, Deephunter: a coverage-guided fuzz testing framework for deep neural
  networks, in: Proceedings of the 28th ACM SIGSOFT International Symposium on
  Software Testing and Analysis, 2019, pp. 146--157.

\bibitem{wodajo2021deepfake}
D.~Wodajo, S.~Atnafu, Deepfake video detection using convolutional vision
  transformer, arXiv preprint arXiv:2102.11126 (2021).

\bibitem{cozzolino2018forensictransfer}
D.~Cozzolino, J.~Thies, A.~R{\"o}ssler, C.~Riess, M.~Nie{\ss}ner, L.~Verdoliva,
  Forensictransfer: Weakly-supervised domain adaptation for forgery detection,
  arXiv preprint arXiv:1812.02510 (2018).

\bibitem{wang2020cnn}
S.-Y. Wang, O.~Wang, R.~Zhang, A.~Owens, A.~A. Efros, Cnn-generated images are
  surprisingly easy to spot... for now, in: Proceedings of the IEEE/CVF
  Conference on Computer Vision and Pattern Recognition, 2020, pp. 8695--8704.

\bibitem{he2016deep}
K.~He, X.~Zhang, S.~Ren, J.~Sun, Deep residual learning for image recognition,
  in: Proceedings of the IEEE conference on computer vision and pattern
  recognition, 2016, pp. 770--778.

\bibitem{li2020face}
L.~Li, J.~Bao, T.~Zhang, H.~Yang, D.~Chen, F.~Wen, B.~Guo, Face x-ray for more
  general face forgery detection, in: Proceedings of the IEEE/CVF Conference on
  Computer Vision and Pattern Recognition, 2020, pp. 5001--5010.

\bibitem{sun2019deep}
K.~Sun, B.~Xiao, D.~Liu, J.~Wang, Deep high-resolution representation learning
  for human pose estimation, in: Proceedings of the IEEE/CVF Conference on
  Computer Vision and Pattern Recognition, 2019, pp. 5693--5703.

\bibitem{sun2019high}
K.~Sun, Y.~Zhao, B.~Jiang, T.~Cheng, B.~Xiao, D.~Liu, Y.~Mu, X.~Wang, W.~Liu,
  J.~Wang, High-resolution representations for labeling pixels and regions,
  arXiv preprint arXiv:1904.04514 (2019).

\bibitem{durall2020watch}
R.~Durall, M.~Keuper, J.~Keuper, Watch your up-convolution: Cnn based
  generative deep neural networks are failing to reproduce spectral
  distributions, in: Proceedings of the IEEE/CVF Conference on Computer Vision
  and Pattern Recognition, 2020, pp. 7890--7899.

\bibitem{frank2020leveraging}
J.~Frank, T.~Eisenhofer, L.~Sch{\"o}nherr, A.~Fischer, D.~Kolossa, T.~Holz,
  Leveraging frequency analysis for deep fake image recognition, in:
  International Conference on Machine Learning, PMLR, 2020, pp. 3247--3258.

\bibitem{durall2019unmasking}
R.~Durall, M.~Keuper, F.-J. Pfreundt, J.~Keuper, Unmasking deepfakes with
  simple features, arXiv preprint arXiv:1911.00686 (2019).

\bibitem{jiang2020focal}
L.~Jiang, B.~Dai, W.~Wu, C.~C. Loy, Focal frequency loss for generative models,
  arXiv preprint arXiv:2012.12821 (2020).

\bibitem{jung2020spectral}
S.~Jung, M.~Keuper, Spectral distribution aware image generation, arXiv
  preprint arXiv:2012.03110 (2020).

\bibitem{huang2020fakepolisher}
Y.~Huang, F.~Juefei-Xu, R.~Wang, Q.~Guo, L.~Ma, X.~Xie, J.~Li, W.~Miao, Y.~Liu,
  G.~Pu, Fakepolisher: Making deepfakes more detection-evasive by shallow
  reconstruction, in: Proceedings of the 28th ACM International Conference on
  Multimedia, 2020, pp. 1217--1226.

\bibitem{wang2018facial}
N.~Wang, X.~Gao, D.~Tao, H.~Yang, X.~Li, Facial feature point detection: A
  comprehensive survey, Neurocomputing 275 (2018) 50--65.

\bibitem{karami2017image}
E.~Karami, S.~Prasad, M.~Shehata, Image matching using sift, surf, brief and
  orb: performance comparison for distorted images, arXiv preprint
  arXiv:1710.02726 (2017).

\bibitem{weickert2016cyclic}
J.~Weickert, S.~Grewenig, C.~Schroers, A.~Bruhn, Cyclic schemes for pde-based
  image analysis, International Journal of Computer Vision 118~(3) (2016)
  275--299.

\bibitem{grewenig2010box}
S.~Grewenig, J.~Weickert, A.~Bruhn, From box filtering to fast explicit
  diffusion, in: Joint Pattern Recognition Symposium, Springer, 2010, pp.
  533--542.

\bibitem{king2009dlib}
D.~E. King, Dlib-ml: A machine learning toolkit, The Journal of Machine
  Learning Research 10 (2009) 1755--1758.

\bibitem{zhang2017automated}
Y.~Zhang, L.~Zheng, V.~L. Thing, Automated face swapping and its detection, in:
  2017 IEEE 2nd International Conference on Signal and Image Processing
  (ICSIP), IEEE, 2017, pp. 15--19.

\bibitem{pedregosa2011scikit}
F.~Pedregosa, G.~Varoquaux, A.~Gramfort, V.~Michel, B.~Thirion, O.~Grisel,
  M.~Blondel, P.~Prettenhofer, R.~Weiss, V.~Dubourg, et~al., Scikit-learn:
  Machine learning in python, the Journal of machine Learning research 12
  (2011) 2825--2830.

\bibitem{matern2019exploiting}
F.~Matern, C.~Riess, M.~Stamminger, Exploiting visual artifacts to expose
  deepfakes and face manipulations, in: 2019 IEEE Winter Applications of
  Computer Vision Workshops (WACVW), IEEE, 2019, pp. 83--92.

\bibitem{nguyen2019multi}
H.~H. Nguyen, F.~Fang, J.~Yamagishi, I.~Echizen, Multi-task learning for
  detecting and segmenting manipulated facial images and videos, in: 2019 IEEE
  10th International Conference on Biometrics Theory, Applications and Systems
  (BTAS), IEEE, 2019, pp. 1--8.

\bibitem{he2015spatial}
K.~He, X.~Zhang, S.~Ren, J.~Sun, Spatial pyramid pooling in deep convolutional
  networks for visual recognition, IEEE transactions on pattern analysis and
  machine intelligence 37~(9) (2015) 1904--1916.

\bibitem{chollet2017xception}
F.~Chollet, Xception: Deep learning with depthwise separable convolutions, in:
  Proceedings of the IEEE conference on computer vision and pattern
  recognition, 2017, pp. 1251--1258.

\bibitem{tan2019efficientnet}
M.~Tan, Q.~Le, Efficientnet: Rethinking model scaling for convolutional neural
  networks, in: International Conference on Machine Learning, PMLR, 2019, pp.
  6105--6114.

\bibitem{Keras}
F.~Chollet, \url{https://keras.io/}. Accessed:2021-03.

\bibitem{qian2020thinking}
Y.~Qian, G.~Yin, L.~Sheng, Z.~Chen, J.~Shao, Thinking in frequency: Face
  forgery detection by mining frequency-aware clues, in: European Conference on
  Computer Vision, Springer, 2020, pp. 86--103.

\bibitem{li2020sharp}
X.~Li, Y.~Lang, Y.~Chen, X.~Mao, Y.~He, S.~Wang, H.~Xue, Q.~Lu, Sharp multiple
  instance learning for deepfake video detection, in: Proceedings of the 28th
  ACM international conference on multimedia, 2020, pp. 1864--1872.

\end{thebibliography}





\end{document}